\documentclass[sigconf]{acmart}

\AtBeginDocument{%
  }

\copyrightyear{2025}
\acmYear{2025}
\setcopyright{cc}
\setcctype{by}
\acmConference[MM '25]{Proceedings of the 33rd ACM International Conference on Multimedia}{October 27--31, 2025}{Dublin, Ireland}
\acmBooktitle{Proceedings of the 33rd ACM International Conference on Multimedia (MM '25), October 27--31, 2025, Dublin, Ireland}\acmDOI{10.1145/3746027.3755244}
\acmISBN{979-8-4007-2035-2/2025/10}

\usepackage{hyperref}
\hypersetup{colorlinks=true,linkcolor=blue,urlcolor=blue}
\usepackage{xcolor}
\usepackage{algorithm}
\usepackage{algpseudocode}
\usepackage{balance}
\usepackage{multirow}
\usepackage{makecell}
\usepackage{colortbl}

\def\ie{{\em i.e.}}
\def\eg{{\em e.g.}}
\newcommand{\shu}{\textcolor{red}}

\usepackage{enumitem}
\begin{document}

\title{Rethinking Individual Fairness in Deepfake Detection}

\author{Aryana Hou}
\authornote{Both authors contributed equally to this research.}
\authornote{Work done as a visiting student under Prof. Shu Hu’s supervision}
\affiliation{%
  \institution{Clarkstown High School South}
  \city{West Nyack}
  \state{New York}
  \country{United States}}
\email{aryanahou@gmail.com}

\author{Li Lin}
\authornotemark[1]
\affiliation{%
  \institution{Purdue University}
  \city{West Lafayette}
  \state{Indiana}
  \country{United States}}
\email{lin1785@purdue.edu}

\author{Justin Li}
\authornotemark[2]
\affiliation{%
  \institution{Carmel High School}
  \city{Carmel}
  \state{Indiana}
  \country{United States}}
\email{justinyli2018@gmail.com}

\author{Shu Hu}
\authornote{Corresponding Author}
\affiliation{%
  \institution{Purdue University}
  \city{West Lafayette}
  \city{Indiana}
  \country{United States}}
\email{hu968@purdue.edu}
\renewcommand{\shortauthors}{Aryana Hou, Li Lin, Justin Li, and Shu Hu}


\begin{abstract}
Generative AI models have substantially improved the realism of synthetic media, yet their misuse through sophisticated DeepFakes poses significant risks. Despite recent advances in deepfake detection, fairness remains inadequately addressed, enabling deepfake markers to exploit biases against specific populations. While previous studies have emphasized group-level fairness, individual fairness (\ie, ensuring similar predictions for similar individuals) remains largely unexplored. In this work, we identify for the \textit{first} time that the original principle of individual fairness fundamentally fails in the context of deepfake detection, revealing a critical gap previously unexplored in the literature. To mitigate it, we propose the \textit{first} generalizable framework that can be integrated into existing deepfake detectors to enhance individual fairness and generalization. Extensive experiments conducted on leading deepfake datasets demonstrate that our approach significantly improves individual fairness while maintaining robust detection performance, outperforming state-of-the-art methods. The code is available at: \url{https://github.com/Purdue-M2/Individual-Fairness-Deepfake-Detection}.
\end{abstract}







\maketitle

\section{Introduction}\label{sec:intro}
\begin{figure}[t]
    \centering
    \includegraphics[width=1.0\linewidth]{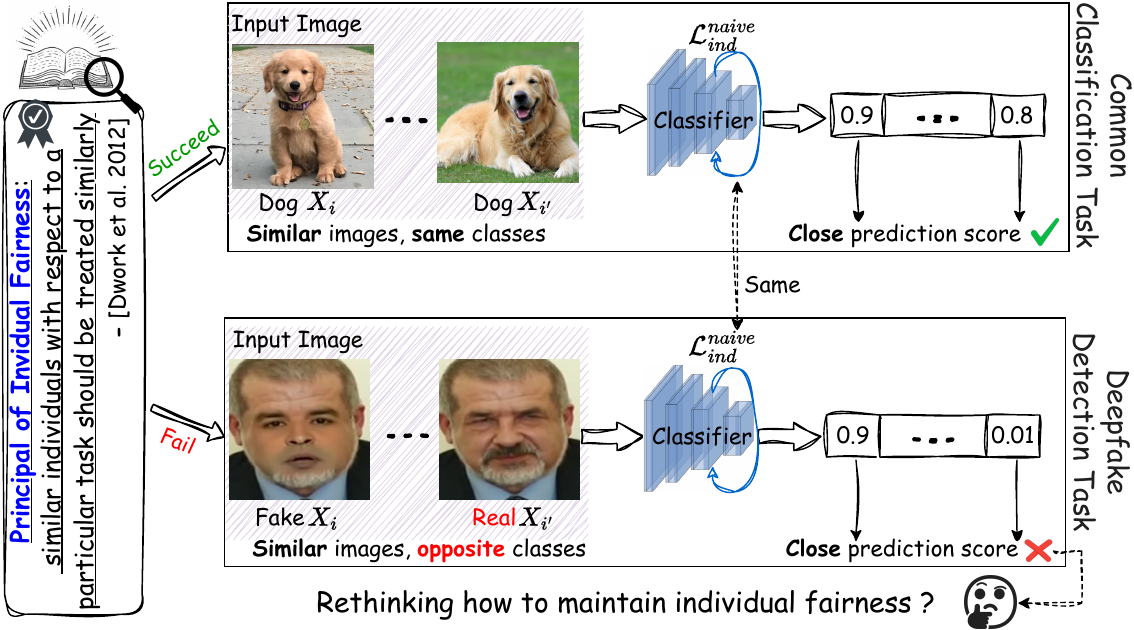}
    \vspace{-8mm}
    \caption{\small \textit{Illustration of the contradiction of individual fairness in deepfake detection. Unlike common classification tasks (Top), where similar images from the same class have close prediction scores satisfying the principle of individual fairness (Left), deepfake detection tasks (Bottom) involve similar images from opposite classes, causing individual fairness to fail.}}
    \vspace{-7mm}
    \label{fig:introduction}
\end{figure}
The past decade has witnessed the \textit{tour de force} of modern generative AI models, such as variational autoencoders~\cite{vahdat2020nvae}, generative adversarial networks~\cite{karras2020analyzing}, and diffusion models~\cite{dhariwal2021diffusion}, driven predominantly by deep neural networks (DNNs). These technologies have significantly advanced the creation of highly realistic images and videos, enriching human experiences and enhancing creative capacities. However, they also carry considerable risks of misuse. One prominent example is ``DeepFakes," which generates highly realistic multimedia content by swapping the faces of individuals without their consent~\cite{westerlund2019emergence}. Deepfake techniques have advanced to a stage where human features and expressions can be replicated with remarkable precision, making it increasingly challenging for ordinary users to differentiate between genuine and fabricated content. The capacity of these technologies to produce persuasive false content capable of manipulating public perceptions poses a serious threat to democratic integrity and individual reputations. Consequently, effective DeepFake detection is critical for combating online misinformation and preserving societal trust in digital information systems.

Recently, numerous deepfake detectors based on DNNs have been developed, aiming to accurately capture spatial and frequency-based artifact characteristics and utilize these learned features to distinguish real from manipulated media~\cite{Sinitsa_2024_WACV, wang2023dire, ma2023exposing, qian2020thinking,krubha2025robust,wang2024spotting,lin2024robust1,zhang2024x,chen2024masked,lin2024detecting,fan2023attacking,zhang2023x,pu2022learning,guo2022open,guo2022robust,wang2023gan}. Beyond pure detection, contemporary research increasingly emphasizes the enhancement of detectors with additional functionalities, including identifying the specific generative model used to create fake media~\cite{guarnera2024level, sha2023fake}, ensuring that detectors maintain generalization capabilities against unseen forgery methods~\cite{yan2024transcending, zheng2025breaking, chen2024diffusionfake, pmlr-v235-chen24ay,ren2024improving,yang2023crossdf}, explaining detection outcomes, localizing manipulated regions within media~\cite{xu2024fakeshield, Hong_2024_CVPR, guo2023hierarchical}, and improving the reliability of detectors even when tested on media of lower quality than training data~\cite{ju2023glff, xu2023exposing, jer2023local, Hooda2024D4}. Despite these advancements, fairness in deepfake detectors remains considerably less explored compared to other aspects.   This oversight significantly limits detectors' applicability, as attackers may exploit biases by targeting particular populations, thus circumventing detection systems.

Only a few studies \cite{wu2025preserving,ju2023improving,lin2024preserving} have developed algorithmic approaches aimed at improving fairness in deepfake detection. For example, \cite{ju2023improving} introduced two fairness losses based on distributionally robust optimization and fairness measures. Both methods encourage similar loss values across demographic groups and demonstrate promising fairness outcomes under intra-domain testing, where the training and test data originate from the same domain. However, these approaches do not ensure fairness generalization in cross-domain evaluation, where test samples are generated from previously unseen forgery methods. To address this limitation, Lin et al.\cite{lin2024preserving} proposed a disentanglement learning framework specifically aimed at enhancing the generalization of fairness for deepfake detectors. \textit{Although both studies effectively improve fairness at the group level, the \textbf{individual-level fairness is largely overlooked}}. 

This individual fairness principle, which asserts that similar individuals should receive similar predictions, was first defined by Dwork et al.~\cite{dwork2012fairness} and has since been widely researched and adopted in machine learning for over a decade. It aims to prevent models from making systematically biased decisions against particular individuals or groups. Unlike traditional fairness approaches, individual fairness does not rely on demographic labels (\eg, race, gender), which are often unavailable, incomplete, or noisy. Thus, individual fairness becomes a valuable demographic-agnostic alternative for mitigating bias.
In the context of deepfake detection, individual fairness is satisfied when the model yields similar predictions for images with similar manipulation characteristics, regardless of their semantic content. Violating individual fairness (two fake or two real images receive inconsistent predictions) indicates that the detection system is unpredictable and untrustworthy, undermining its utility in high-stakes applications such as forensic analysis or media verification. However, according to individual fairness, since deepfake images closely resemble real images, the detector should assign the fake image the same label as the real target image, directly contradicting the primary goal of deepfake detection (see Fig. \ref{fig:introduction}). \textit{This critical issue has yet to be explored in recent deepfake detection literature, and the specific scenarios where individual fairness fails have not been thoroughly examined even within broader machine learning fields.} Such a gap prompts two essential research questions: How can individual fairness be effectively restored in deepfake detection models? And how can we ensure that the individual fairness properties of trained models generalize reliably when deployed in unseen scenarios?

To answer the above questions, we propose the \textbf{first} general framework designed to integrate seamlessly into most existing deepfake detection methods to improve their individual fairness. Specifically, we first empirically demonstrate the failure of the individual fairness principle on deepfake datasets and further illustrate that merely transforming original images into their frequency representations does not restore individual fairness. After rethinking the conventional individual fairness loss formulation, we develop a novel learning objective by leveraging anchor-based learning and introducing a new procedure aimed at explicitly exposing forgery-related feature representations for similarity calculation. Training with our proposed learning objective on a flattened loss landscape enhances individual fairness and generalization capability of detectors, while also improving detection performance. Our main contributions are as follows: 
\begin{enumerate}[noitemsep,topsep=0pt,leftmargin=*]
    \item We identify, for the \textbf{first} time, the fundamental failure of the individual fairness principle in the context of deepfake detection, highlighting a critical gap previously unexplored.
    \item We propose the \textbf{first} general framework designed to seamlessly integrate into a wide range of existing deepfake detection methods, effectively enhancing their individual fairness as well as generalization capabilities.
    \item Extensive experimental evaluations on several leading deepfake datasets demonstrate that our proposed approach outperforms state-of-the-art methods in improving individual fairness and detection utility simultaneously.
\end{enumerate}

\section{Related Work}
\smallskip
\noindent
\textbf{Individual Fairness. }
To the best of our knowledge, individual fairness in deepfake detection remains unexplored. However, this challenge has been studied in broader machine learning contexts.
The concept of individual fairness was first introduced by Dwork et al.~\cite{dwork2012fairness} under the principle that 
``\textit{similar individuals with respect to a particular task should be treated similarly}.'' 
The major challenge in individual fairness is the definition and estimation of an appropriate similarity metric, which remains a bottleneck in its broader application~\cite{mukherjee2020two}. Several recent studies have extended individual fairness beyond its original definition. Kearns et al.~\cite{sharifi2019average} proposed average individual fairness, which balances statistical and individual fairness by ensuring fairness across both individuals and classification tasks. Mukherjee et al.~\cite{mukherjee2020two} presented two data-driven approaches to learn fair metrics, making individual fairness more practical to broader machine learning tasks. However, \cite{fleisher2021s} argued that similarity-based fairness can still encode human biases, leading to systematic disparities if the metric itself is flawed. Furthermore, unlike traditional image classification, where samples of the same class exhibit higher similarity based on observed semantics, deepfakes often exhibit greater similarity across different classes, as demonstrated in our following Motivation section. Consequently, \textit{the conventional principle of individual fairness is not directly applicable, requiring us to rethink individual fairness in deepfake detection}.

\vspace{-1mm}

\begin{figure*}[t]
    \centering
    \includegraphics[width=0.95\linewidth]{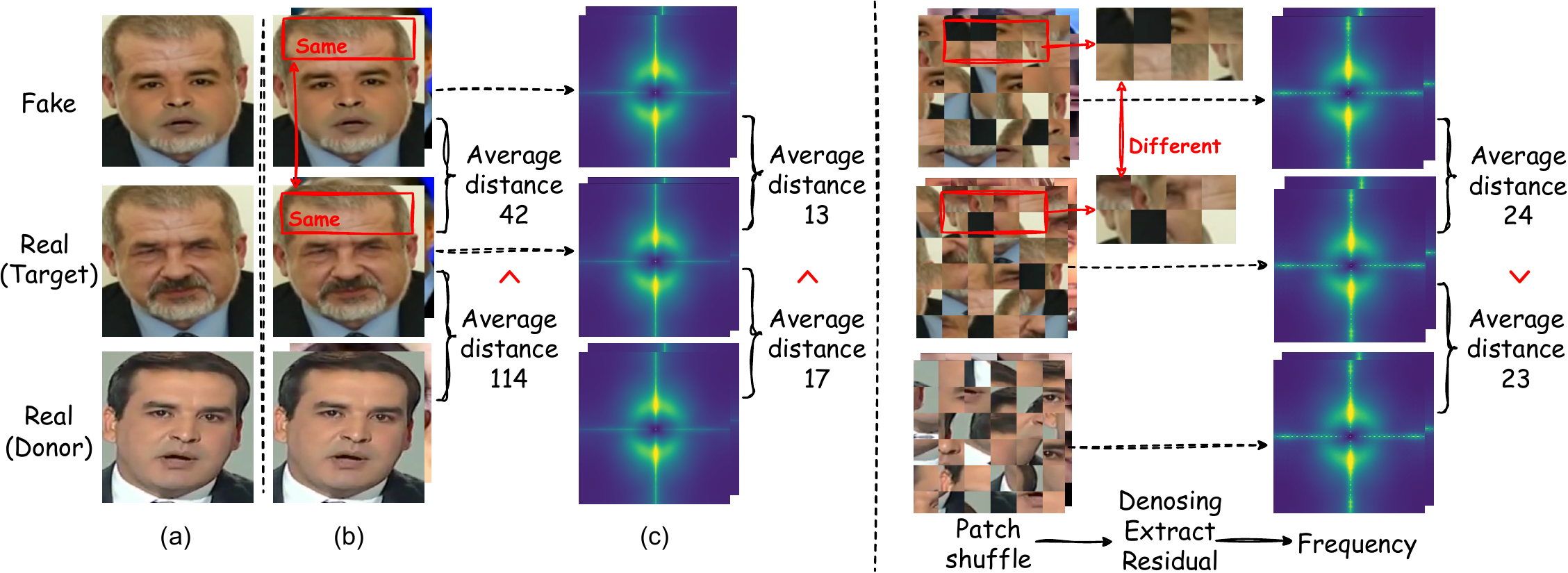}
    \vspace{-2mm}
    \caption{\small \textbf{\textit{(Left) Illustration of the failure of individual fairness in deepfake detection. (a) A fake image is created by manipulating a real target face with facial regions from a real donor. (b) Pixel-level distances between 1,000 averaged images reveal that fake images are closer to their real targets than the real targets are to donors. (c) Frequency representation fails in removing semantic similarity. (Right) The success of our proposed semantic-agnostic individual fairness.}}}
    \vspace{-2mm}
    \label{fig:motivation}
\end{figure*}

\section{Motivation}\label{sec:motivation}

\subsection{The ``Failure'' of Individual Fairness}


The principle of individual fairness usually states that similar individuals should receive similar predictions. However, applying this principle in deepfake detection may lead to inherent contradictions. This is because deepfake images are created by manipulating a real \textbf{target} face using features from another real face, known as the \textbf{donor}, as shown in Fig.~\ref{fig:motivation} (a). Consequently, the resulting \textbf{fake} image retains significant semantic and identity-related attributes of the target, while also incorporating certain characteristics of the donor.  Therefore, we guess that the difference between the fake and real (target) images could be smaller than that between the real (target) and real (donor) images.

To verify this hypothesis, we measure the Euclidean distance between fake images and their corresponding real target images, as well as between real target and real donor images. For clarity, we use variables \(X^f\), \(X^t\), and \(X^d\) to represent fake, real target, and real donor images, respectively. Specifically, we randomly sample $I=1000$ triplets from the dataset, where each triplet consists of a fake image ($X_i^f$), a real target image ($X_i^t$), and a real donor image ($X_i^d$) for $i=1, 2, \ldots, I$. Before distance calculation, we first normalize all images as a pre-processing step. Then, for each triplet, we compute the Euclidean distance: $d(X_i^f, X_i^t) = \|X_i^f - X_i^t\|_2$ and $d(X_i^t, X_i^d) = \|X_i^t - X_i^d\|_2$. Then, we calculate the average distances across all triplets as $\bar{d}(X^f, X^t) = \frac{1}{I} \sum_{i=1}^I d(X_i^f, X_i^t)$ and $\bar{d}(X^t, X^d) = \frac{1}{I} \sum_{i=1}^I d(X_i^t, X_i^d)$. As illustrated in Fig.~\ref{fig:motivation} (b), the distance $\bar{d}(X^f, X^t)=42$ is significantly smaller than $\bar{d}(X^t,X^d)=114$, supporting our hypothesis.


\subsection{Does Frequency Representation Work?} 
If individual fairness constraint is naively incorporated into the loss function, it would force fake and real (target) to produce similar outputs, thereby reducing the separability between fake and real images. This directly contradicts the core objective of deepfake detection, which is to maximize the distinction between fake and real images rather than minimize it.
In fact, deepfake detection requires the model to focus not on high-level semantic similarity but on low-level manipulations and imperceptible artifacts that differentiate fake images from real ones. Therefore, an effective solution must aim to remove as much semantic information as possible to meet the goal of deepfake detection while improving individual fairness.

A straightforward approach to mitigating the impact of semantic features is to transform images into the frequency domain, where high-frequency components are expected to better capture manipulation artifacts. Specifically, we first convert each randomly sampled RGB image $X_i^{z}$ into the frequency domain using the discrete Fourier transform and compute the corresponding power spectrum $ P_{i}^z=|\mathcal{F}(X_i^z)|^2$ for $z \in \{f, t, d\}, i=1,\dots,I$. Then, for each triplet, we compute the Euclidean distances between $P_{i}^z$: $d(P_i^f, P_i^t)=\|P_i^f - P_i^t\|_2$ and $d(P_i^t, P_i^d) = \|P_i^t - P_i^d\|_2$. Then, we calculate the average distances across all triplets as $\bar{d}(P^f, P^t) = \frac{1}{I} \sum_{i=1}^I d(P_i^f, P_i^t)$ and $\bar{d}(P^t, P^d) = \frac{1}{I} \sum_{i=1}^I d(P_i^t, P_i^d)$.
As shown in Fig.~\ref{fig:motivation} (c), even in the frequency domain, the distance $\bar{d}(P^f, P^t)=13$ remains smaller than $\bar{d}(P^t, P^d)=17$, suggesting that direct utilization of frequency-domain representations is ineffective. One potential reason could be that the semantic information is still embedded in the frequency representation. Indeed, \cite{li2024freqblender} has shown that frequency features still retain some high-level semantic information. 
Therefore, simply converting images from RGB space to frequency space is insufficient to resolve individual fairness issues in deepfake detection. 

\section{The Naive Approach}

Given a training dataset $\mathcal{S}=\{(X_i, Y_i)\}_{i=1}^n$ with size $n$, where $X \in \mathbb{R}^{d}$ is the image with dimension $d$, $Y \in \{0,1\}$ is the corresponding label (0 for fake, 1 for real). 
A straightforward method to enhance individual fairness in deepfake detection is to incorporate a fairness regularization term into a baseline classifier. Specifically, let us consider a deep learning model with a feature extractor $E(\cdot)$ and a classifier head $h(\cdot)$. The conventional training for a deepfake detector is to minimize the following learning objective:
$\mathcal{L}_{CE} = \frac{1}{n}\sum_{i=1}^{n} C(h(E(X_i)),Y_i)$, 
where $C(\cdot,\cdot)$ represents the Cross-Entropy (CE) loss.  To enforce individual fairness, a naive approach widely used in existing works \cite{ju2023improving, lin2024preserving} is to introduce an additional fairness regularization term, $\mathcal{L}^{naive}_{ind}$, which penalizes discrepancies in model outputs for similar input images. The resulting objective function can be represented as:
\begin{equation}
    \begin{aligned}
    &\mathcal{L}_{naive} = \mathcal{L}_{CE} + \lambda\mathcal{L}^{naive}_{ind},\\
     & \text{where} \ \mathcal{L}^{naive}_{ind} = \sum_{i=1}^{n-1} \sum_{j=i+1}^{n} \big[\underbrace{|h(E(X_i)) - h(E(X_{j}))|}_{\spadesuit} - \tau\underbrace{\left\|X_i - X_{j}\right\|_2}_{\clubsuit}\big]_+.
    \end{aligned}
\label{eq:base}
\end{equation}
Here $[\cdot]_{+}$ is the hinge function, defined as $[a]_{+} = \max\{0, a\}$ for any $a \in \mathbb{R}$, ensuring non-negative penalties. The term $\|\cdot\|_2$ is the $\ell_2$ norm, measuring the similarity between input images. The hyperparameter $\lambda \in \mathbb{R}$ balances detection accuracy and individual fairness, while $\tau \in \mathbb{R}$ is a predefined scale factor. 

Despite its intuitive formulation, this approach presents two fundamental issues:
\textbf{(1)} It enforces similarity based on raw image-space distances (see the term $\clubsuit$), which primarily capture high-level semantic attributes rather than fine-grained manipulation artifacts. Consequently, deepfake images that retain substantial semantic similarity to their real counterparts may yield similar predictions, thus impairing the model’s ability to distinguish between fake and real images. This contradiction underscores the necessity for a more targeted approach to enforcing individual fairness.
\textbf{(2)} Even if individual fairness is maintained successfully during training, fairness performance may not generalize effectively when the detector encounters scenarios different from the training domain - a challenge referred to as the generalization issue in cross-domain testing. This limitation arises because the $\mathcal{L}_{CE}$ loss and the term $\spadesuit$ operate exclusively on the feature representations of the training images, which lacks forgery domain diversity. To address these two issues, we propose a novel method to enhance the detector’s individual fairness while preserving its generalization capability.

\begin{figure*}[t]
    \centering
    \includegraphics[width=1.0\linewidth]{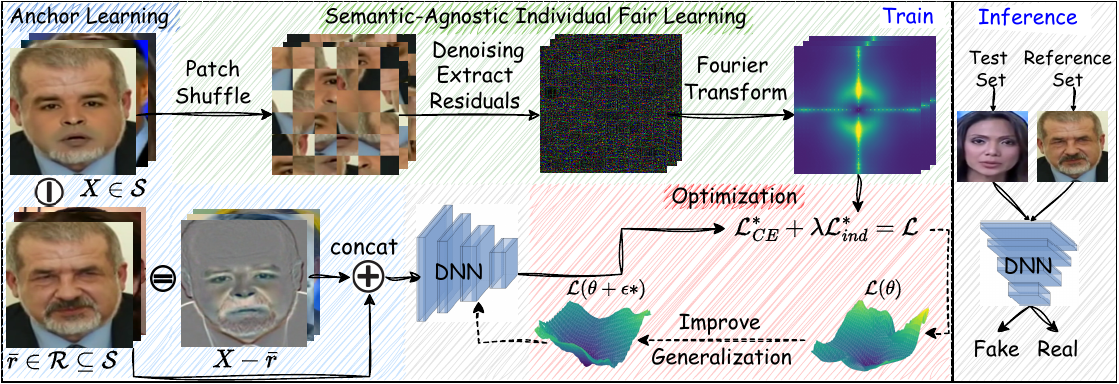}
    \vspace{-6mm}
    \caption{\small \textbf{\textit{Overview of our proposed method. 1) Anchor Learning first transforms each input image into a residual representation relative to a randomly selected reference, enhancing the generalization of individual fairness. 2) Semantic-Agnostic Individual Fair Learning further mitigates semantic bias by patch shuffling, residual extraction, and frequency-domain transformation to break the semantic correlations and expose forgery artifacts. 3) For the optimization, we flatten the loss landscape to further enhance fairness generalization.}}}
    \vspace{-1mm}
    \label{fig:framework}
\end{figure*}

\section{The Proposed Method}
An overview of our framework is shown in Fig. \ref{fig:framework}.
\subsection{Anchor Learning}
First, we address the issue arising from the $\mathcal{L}_{CE}$ loss and the term $\spadesuit$. While existing approaches~\cite{lin2024preserving, ren2024improving, Yan_2023_ICCV, yan2024transcending}, such as disentanglement learning and data augmentation, have been proposed to enhance generalization in deepfake detection, these methods typically result in increased model complexity or larger training datasets. Consequently, they may not be applicable to integrate into advanced detection methods or restricted training environments. This motivates us to develop a more efficient and broadly generalizable approach.

Inspired by \cite{NEURIPS2024_ad405c24}, which introduced anchor training to enhance generalization to unseen domains, we leverage this approach to achieve our goal. Specifically, anchor training reparameterizes each input image  $X$ into a tuple $[\tilde{r}, X-\tilde{r}]$, where $\tilde{r} \sim P_r$ is the reference sample drawn from a reference distribution $P_r$, and $X-\tilde{r}$ is the residual capturing the difference between the input and the reference. This encourages the model to focus on manipulation-specific features rather than domain or identity-dependent semantics, thus suppressing spurious correlations and helping the model generalize better across unseen domains. The tuple is concatenated along the channel axis and fed into the deepfake detection model for training. The model is thus trained to learn the joint distribution $P_{(r, \Delta)}$, where $\Delta$ represents the residual distribution, ensuring that the prediction for a given input $X$ is consistent regardless of the choice of reference $\tilde{r}$. 

In practice, for each input image $X_i$, we sample a reference $\bar{r}\in \mathcal{R}$, where a reference set $\mathcal{R} \subseteq \mathcal{S}$ is sampled from the training dataset $\mathcal{S}$ while $X_i\notin \mathcal{R}$. Then, we compute the residual $d_i=X_i-\bar{r}$ and form the concatenated anchor input $X_{i}^{a}=concat([\bar{r},d_i])$. 
To further discourage models from learning shortcuts based solely on residual information, we employ reference masking~\cite{NEURIPS2024_ad405c24}. Specifically, with probability $\alpha$, the reference component of the anchor input is masked (set to zero): $X_{i}^{a}=concat([0,d_i])$. 
During training, when the reference is masked, we encourage the model to produce high-entropy (uniform) predictions by minimizing the cross-entropy loss between the model’s output and a uniform prior $\mathcal{P}$ over two classes (\ie, the probability of any class is 1/2). 
This ensures that the reference $\bar{r}$ plays a meaningful role in the decision-making process and mitigates the risk of learning non-generalizable shortcuts. To sum up, the original $\mathcal{L}_{CE}$ loss is replaced with:
\begin{equation}\small
\begin{aligned}
        &\mathcal{L}_{CE}^* = 
        \frac{1}{n} \sum_{i=1}^{n} \big[(1 - \mathbb{I}_{\text{mask}})\cdot C(h(E(X_{i}^a)), Y_i)+ \mathbb{I}_{\text{mask}} \cdot C(h(E((X_{i}^{a})), \mathcal{P})\big],
\end{aligned}
\label{eq:ce_loss}
\end{equation}
where $\mathbb{I}_{\text{mask}}$ is an indicator function that takes the value 1 with probability $\alpha$ (applying the mask) and 0 otherwise. 

Similarly, the term $\spadesuit$ will be replaced with $|h(E(X_i^a))-h(E(X_{j}^a))|$. To avoid situations where the mask applies only to a single sample (\ie, rendering prediction comparisons less meaningful), we implement masking at the batch level in practice. This ensures that both $X_i^a$ and $X_j^a$ share an identical mask configuration. Note that although we utilize the existing anchor training approach from~\cite{NEURIPS2024_ad405c24}, this is the first time it has been explored within deepfake detection. Furthermore, we adapt it specifically to enhance the generalization of individual fairness, which has not previously been investigated.

\subsection{Semantic-Agnostic Individually Fair Learning}
Second, we propose a new approach to accurately assess similarities between individual samples to address the issue arising from the $\clubsuit$ term. Our approach reduces the dominant influence of semantic information, focusing instead on common cues related to manipulation techniques derived from exposed artifacts. To do so, we have to remove semantic biases at the image level, including high-level concepts such as identity, facial attributes, and background context, which are often entangled with model predictions. 

\smallskip
\noindent
\textbf{Breaking Semantic Correlations}.
Since the high similarity in semantic features across different classes in deepfake datasets (as shown in Fig. \shu{1} (a) and (b)) conflicts with the definition of individual fairness loss (see Eq. (\ref{eq:base})), training a detector with the above naive approach fails to mitigate bias. This bias further arises as models overfit identity-related attributes rather than focus on the manipulation artifacts. To address this challenge, we employ a patch shuffle strategy.
First, given an input image \( X \), we divide it into multiple non-overlapping patches of size \( P \times P \):
$
\{p_o\} = \text{Partition}(X, P),\quad o = 1, 2, \dots, K \times K
$
where \( K \times K \) represents the total number of patches obtained after segmentation. To disrupt global semantic dependencies while preserving local details, we randomly shuffle these patches into a new sequence $\{p_m\}, (m = 1, 2, \dots, K \times K)$. Finally, we reconstruct the image by rearranging the shuffled patches back to their original dimensions, forming \( X' \), which serves as input for subsequent processing. This transformation ensures that the model focuses on fine-grained manipulation artifacts rather than identity-related semantic structures.

\smallskip
\noindent
\textbf{Exposing Artifacts}.
Generative models typically introduce subtle but systematic artifacts during the synthesis process. These artifacts often remain concealed under high-level semantic details, making them difficult to discern when analyzing raw images directly. To isolate and amplify these telltale signs, we adopt a residual-based strategy~\cite{corvi2023intriguing}.  Specifically, after applying patch shuffling to an input image $X$, we generate a denoised version $D(X')$ with a filter from~\cite{zhang2017beyond} to capture the smooth, semantic content. We then compute the residual $\overline{X}'$ by subtracting the denoised result from $X'$:
$\overline{X}' = X' - D(X')$. 
Crucially, we compute the residual after patch shuffling, thereby exposing the remaining synthesized artifacts once semantic connections are disrupted.

\smallskip
\noindent
\textbf{Viewing from Frequency Domain}.
While residuals capture spatial irregularities indicative of manipulation, analyzing the frequency domain can reveal additional clues linked to generative pipelines~\cite{corvi2023intriguing, corvi2023detection, wang2020cnn}. Natural images acquired through physical cameras typically exhibit characteristic power-law decays in their Fourier spectra and anisotropic patterns shaped by lenses and sensors. By contrast, AI-generated or manipulated images often leave checkerboard signatures or spectral peaks due to transposed convolutions or latent up-sampling. Consequently, inspecting residuals in the frequency domain offers a complementary view that can confirm or refine forgery indicators observed in the spatial domain.
\textit{Most importantly, this process reveals similar patterns between samples originating from the same source (\eg, real or the same forgery method). This is particularly useful for determining whether two samples are similar based solely on generation or manipulation technique types.} 
We consider the discrete Fourier transform
\(\mathcal{F}\{\cdot\}\) of each residual image $\overline{X}'$ by getting its frequency domain representation $\mathcal{F}(\overline{X}')$.

After a series of processing steps—including patch shuffle, denoising, residual extraction, and Fourier transform—we compute the average distance of the resulting representations using the same settings as of the experiment in the Motivation Section \ref{sec:motivation}. As shown in the right part of Fig.~\ref{fig:motivation}, our method effectively separates the fake and the real, bringing the fake–real target distance farther than the real target–real donor distance.

\smallskip
\noindent
\textbf{New Individual Fairness Loss}.
Based on the above processing, we have successfully broken high-level semantical features and exposed the artifacts. Therefore, we can design the new individual loss as follows: 
\begin{equation}
\mathcal{L}^{*}_{ind} = \sum_{i=1}^{n-1} \sum_{j=i+1}^{n} \big[|h(E(X_i^a)) - h(E(X_{j}^a))| - \tau\|\mathcal{F}(\overline{X}'_i) - \mathcal{F}(\overline{X}'_{j})\|_2\big]_+.
\label{eq:ours_ind}
\end{equation}

\subsection{Optimization}
Lastly, we optimize the following final learning objective formulated as a weighted combination of $\mathcal{L}_{CE}^*$ and $\mathcal{L}_{ind}^*$:
\begin{equation}
    \mathcal{L}= \mathcal{L}_{CE}^* + \lambda\mathcal{L}_{ind}^*.
\label{eq:final_loss}
\end{equation}
To improve generalization and avoid suboptimal solutions common in overparameterized DNNs, we apply the sharpness-aware minimization (SAM) technique \cite{foret2020sharpness} to flatten the loss landscape, which has been proven to efficiently improve generalization~\cite{lin2024preserving, ren2024improving, lin2024robust}.
Specifically, denoting the model weights of the whole framework as $\theta$, this involves finding an optimal $\epsilon^*$ to perturb $\theta$ in a way that maximizes the loss, expressed as:
\begin{equation}
    \begin{aligned}        
    \epsilon^*=\arg\max_{\|\epsilon\|_2\leq \gamma} \mathcal{L}(\theta+\epsilon)\approx\arg\max_{\|\epsilon\|_2\leq \gamma} \epsilon^\top\nabla_\theta \mathcal{L}=\gamma \frac{\nabla_\theta \mathcal{L}}{\|\nabla_\theta \mathcal{L}\|_2},
    \end{aligned}
\label{eq:eq_sam}
\end{equation} 
Here, $\gamma$ is a hyperparameter that controls the perturbation magnitude, and $\nabla_\theta \mathcal{L}$ is the gradient of $\mathcal{L}$ with respect to $\theta$. The approximation term is derived using a first-order Taylor expansion, assuming $\epsilon$ is small. The final equation is obtained by solving a dual norm problem. Thus, the model weights are updated by solving: $
    \begin{aligned}      
    \min_\theta \mathcal{L}\textbf{(}\theta+\epsilon^*\textbf{)}
    \end{aligned}$.
The underlying idea is that perturbing the model in the direction of the gradient norm increases the loss value, thereby improving generalization. We optimize the SAM objective using stochastic gradient descent, as detailed in Algorithm 1 of the Appendix. 

During inference, we randomly select a single reference from the predefined reference set and evaluate on the test dataset, following the same inference protocol from \cite{NEURIPS2024_ad405c24}.

\section{Experiment}
\begin{table*}[t!]
\centering
\scalebox{0.77}{
\begin{tabular}{c|ccc|ccc|ccc|ccc|ccc}
\hline
                         & \multicolumn{3}{c|}{FF++}                                                                     & \multicolumn{3}{c|}{DFDC}                                                                     & \multicolumn{3}{c|}{Celeb-DF}                                                                 & \multicolumn{3}{c|}{DFD}                                                                      & \multicolumn{3}{c}{AI-Face}                                                                                       \\ \cline{2-16} 
                         & Utility(\%)\textuparrow                 & \multicolumn{2}{c|}{Fairness\textdownarrow}                                   & Utility(\%)\textuparrow                 & \multicolumn{2}{c|}{Fairness\textdownarrow}                                   & Utility(\%)\textuparrow                 & \multicolumn{2}{c|}{Fairness\textdownarrow}                                   & Utility(\%)\textuparrow                 & \multicolumn{2}{c|}{Fairness\textdownarrow}                                   & Utility(\%)\textuparrow                   & \multicolumn{2}{c}{Fairness\textdownarrow}                                                      \\ \cline{2-16} 
\multirow{-3}{*}{Method} & AUC                         &$\mathcal{L}_{ind}^{naive}$      &$\mathcal{L}_{ind}^{*}$      & AUC                         &$\mathcal{L}_{ind}^{naive}$      &$\mathcal{L}_{ind}^{*}$      & AUC                         &$\mathcal{L}_{ind}^{naive}$      &$\mathcal{L}_{ind}^{*}$      & AUC                         &$\mathcal{L}_{ind}^{naive}$      &$\mathcal{L}_{ind}^{*}$      & AUC                           &$\mathcal{L}_{ind}^{naive}$               &$\mathcal{L}_{ind}^{*}$               \\ \hline \hline
Ori                      & 92.76                       & 1.5928                         & 1.2472                         & 92.39                       & 2.1227                         & 1.7047                         & 97.00                       & 1.8844                         & 1.5162                         & 92.94                       & 2.1144                         & 1.7138                         & \textbf{99.77}                & 6.6605                                  & 6.1603                                  \\
Naive                    & 93.12                       & 1.3977                         & 1.1082                         &  91.45                           &        1.9779                        & 1.5080                               & 97.20                       & 1.4277                         & 1.1132                         & 93.84                       & 1.5552                         & 1.2164                         & 98.98                         & 4.7780                                  & 4.2831                                  \\
\rowcolor[HTML]{EFEFEF} 
\textbf{Ours}            & \textbf{98.53}              & \textbf{0.8153}                & \textbf{0.4709}                & \textbf{96.89}              & \textbf{1.1107}                & \textbf{0.7452}                & \textbf{99.15}              & \textbf{0.7822}                & \textbf{0.4721}                & \textbf{96.23}              & \textbf{1.1285}                & \textbf{0.7419}                & 99.20                         & \cellcolor[HTML]{EFEFEF}\textbf{4.5609} & \cellcolor[HTML]{EFEFEF}\textbf{4.0861} \\
\color[HTML]{036400} Diff.                    & {\color[HTML]{036400} +5.77} & {\color[HTML]{036400} -0.7775} & {\color[HTML]{036400} -0.7763} & {\color[HTML]{036400} +4.50} & {\color[HTML]{036400} -1.0120} & {\color[HTML]{036400} -0.9595} & {\color[HTML]{036400} +2.15} & {\color[HTML]{036400} -1.1022} & {\color[HTML]{036400} -1.0441} & {\color[HTML]{036400} +3.29} & {\color[HTML]{036400} -0.9859} & {\color[HTML]{036400} -0.9719} & {\color[HTML]{036400} -00.57} & {\color[HTML]{036400} -2.0996}          & {\color[HTML]{036400} -2.0742}          \\ \hline
\end{tabular}
}
\caption{\small \textit{{Intra-domain evaluation results across five datasets compared to Ori and Naive. All method are based on Xception backbone. We provide the difference (Diff.) between our proposed and the Ori on each dataset with {\color[HTML]{036400} green}. \textuparrow means higher is better and \textdownarrow means lower is better.} }}
\vspace{-5mm}
\label{tab:in_domain}
\end{table*}

\begin{table*}[t!]
\centering
\scalebox{0.84}{
\begin{tabular}{c|c|ccc|ccc|ccc|ccc}
\hline
                                  &                                       & \multicolumn{3}{c|}{FF++}                                                                                                  & \multicolumn{3}{c|}{DFDC}                                                                                                  & \multicolumn{3}{c|}{Celeb-DF}                                                                                              & \multicolumn{3}{c}{DFD}                                                                           \\ \cline{3-14} 
                                  &                                       & Utility(\%)\textuparrow                            & \multicolumn{2}{c|}{Fairness\textdownarrow}                                                     & Utility(\%)\textuparrow                            & \multicolumn{2}{c|}{Fairness\textdownarrow}                                                     & Utility(\%)\textuparrow                            & \multicolumn{2}{c|}{Fairness\textdownarrow}                                                     & Utility(\%)\textuparrow                            & \multicolumn{2}{c}{Fairness\textdownarrow}                                          \\ \cline{3-14} 
\multirow{-3}{*}{Backbone}        & \multirow{-3}{*}{Method}              & AUC                                    &$\mathcal{L}_{ind}^{naive}$               &$\mathcal{L}_{ind}^{*}$               & AUC                                    &$\mathcal{L}_{ind}^{naive}$               &$\mathcal{L}_{ind}^{*}$               & AUC                                    &$\mathcal{L}_{ind}^{naive}$               &$\mathcal{L}_{ind}^{*}$               & AUC                                    &$\mathcal{L}_{ind}^{naive}$               &$\mathcal{L}_{ind}^{*}$               \\ \hline \hline
                                  & Ori                                   & 92.76                                  & 1.5928                                  & 1.2472                                  & 58.81                                  & 3.4204                                  & 2.9289                                  & 62.65                                  & 1.6279                                  & 1.2760                                  & 68.20                                  & 2.1728                                  & 1.7757          \\
                                  & Naive                                 & 93.12                                  & 1.3977                                  & 1.1082                                  & 57.71                                  & 2.3442                                  & 1.9277                                  & 71.76                                  & 1.7519                                  & 1.4132                                  & 64.77                                  & 2.2195                                  & 1.8446                      \\
                                  & \cellcolor[HTML]{EFEFEF}\textbf{Ours} & \cellcolor[HTML]{EFEFEF}\textbf{98.53} & \cellcolor[HTML]{EFEFEF}\textbf{0.8153} & \cellcolor[HTML]{EFEFEF}\textbf{0.4709} & \cellcolor[HTML]{EFEFEF}\textbf{62.06} & \cellcolor[HTML]{EFEFEF}\textbf{0.8063} & \cellcolor[HTML]{EFEFEF}\textbf{0.8054} & \cellcolor[HTML]{EFEFEF}\textbf{76.62} & \cellcolor[HTML]{EFEFEF}\textbf{1.1957} & \cellcolor[HTML]{EFEFEF}\textbf{0.8221} & \cellcolor[HTML]{EFEFEF}\textbf{80.76} & \cellcolor[HTML]{EFEFEF}\textbf{1.2699} & \cellcolor[HTML]{EFEFEF}\textbf{0.8791} \\
\multirow{-4}{*}{Xception}        & {\color[HTML]{036400} Diff.}          & {\color[HTML]{036400} +5.77}           & {\color[HTML]{036400} -0.7775}          & {\color[HTML]{036400} -0.7763}          & {\color[HTML]{036400} +3.25}           & {\color[HTML]{036400} -2.6141}          & {\color[HTML]{036400} -2.1235}          & {\color[HTML]{036400} +13.97}          & {\color[HTML]{036400} -0.4322}          & {\color[HTML]{036400} -0.4539}          & {\color[HTML]{036400} +12.56}          & {\color[HTML]{036400} -0.9029}          & {\color[HTML]{036400} -0.8966}   \\ \hline
                                  & Ori                                   & 94.32                                  & 2.3504                                  & 1.8718                                  & 59.49                                  & 3.6806                                  & 3.1626                                  & 65.61                                  & 3.3638                                  & 2.8932                                  & 71.86                                  & 3.0166                                  & 2.5175                                \\
                                  & Naive                                 & 95.71                                  & 1.3807                                  & 1.0757                                  & 59.66                                  & 2.6222                                  & 2.2041                                  & 77.96                                  & 2.1250                                  & 1.7724                                  & 71.21                                  & 1.9494                                  & 0.9494                                 \\
                                  & \cellcolor[HTML]{EFEFEF}\textbf{Ours} & \cellcolor[HTML]{EFEFEF}\textbf{96.89} & \cellcolor[HTML]{EFEFEF}\textbf{0.7941} & \cellcolor[HTML]{EFEFEF}\textbf{0.4508} & \cellcolor[HTML]{EFEFEF}\textbf{62.49} & \cellcolor[HTML]{EFEFEF}\textbf{1.0033} & \cellcolor[HTML]{EFEFEF}\textbf{0.6355} & \cellcolor[HTML]{EFEFEF}\textbf{80.64} & \cellcolor[HTML]{EFEFEF}\textbf{0.8213} & \cellcolor[HTML]{EFEFEF}\textbf{0.4841} & \cellcolor[HTML]{EFEFEF}\textbf{79.27} & \cellcolor[HTML]{EFEFEF}\textbf{1.1173} & \cellcolor[HTML]{EFEFEF}\textbf{0.7377} \\
\multirow{-4}{*}{ResNet-50}       & {\color[HTML]{036400} Diff.}          & {\color[HTML]{036400} +2.57}           & {\color[HTML]{036400} -1.5563}          & {\color[HTML]{036400} -1.4210}          & {\color[HTML]{036400} +3.00}           & {\color[HTML]{036400} -2.6773}          & {\color[HTML]{036400} -2.5271}          & {\color[HTML]{036400} +15.03}          & {\color[HTML]{036400} -2.5425}          & {\color[HTML]{036400} -2.4091}          & {\color[HTML]{036400} +7.41}           & {\color[HTML]{036400} -1.8993}          & {\color[HTML]{036400} -1.7798}          \\ \hline
                                  & Ori                                   & 95.92                                  & 1.7932                                  & 1.4577                                  & 57.81                       & 5.4014                                  & 4.9215                                  & 62.35                                  & 1.3927                                  & 1.0483                                  & 75.54                                  & 2.6022                                  & 2.1948                                 \\
                                  & Naive                                 & 96.79                                  & 1.2498                                                                             & 0.8575                                                         & 55.94                                                & 1.9404                                                         & 1.4838                                                         & 63.38                                                         & 1.0590                                                         & 0.6686                                                         & 78.84                                                         & 1.9728                                                         & 1.5120                                                         \\
                                  & \cellcolor[HTML]{EFEFEF}\textbf{Ours} & \cellcolor[HTML]{EFEFEF}\textbf{97.88} & \multicolumn{1}{r}{\cellcolor[HTML]{EFEFEF}\textbf{0.7907}} & \cellcolor[HTML]{EFEFEF}\textbf{0.3940} & \cellcolor[HTML]{EFEFEF}\textbf{59.44}          & \cellcolor[HTML]{EFEFEF}\textbf{0.6943} & \cellcolor[HTML]{EFEFEF}\textbf{0.3403} & \cellcolor[HTML]{EFEFEF}\textbf{68.56} & \cellcolor[HTML]{EFEFEF}\textbf{0.4827} & \cellcolor[HTML]{EFEFEF}\textbf{0.1521} & \cellcolor[HTML]{EFEFEF}\textbf{83.14} & \cellcolor[HTML]{EFEFEF}\textbf{0.9088} & \cellcolor[HTML]{EFEFEF}\textbf{0.4896} \\
\multirow{-4}{*}{EfficientNet-B3} & {\color[HTML]{036400} Diff.}          & {\color[HTML]{036400} +1.96}           & {\color[HTML]{036400} -1.0025}                              & {\color[HTML]{036400} -1.0637}          & {\color[HTML]{036400} +1.63}           & {\color[HTML]{036400} -4.7071}          & {\color[HTML]{036400} -4.5812}          & {\color[HTML]{036400} +6.21}           & {\color[HTML]{036400} -0.9100}          & {\color[HTML]{036400} -0.8962}          & {\color[HTML]{036400} +7.60}           & {\color[HTML]{036400} -1.6934}          & {\color[HTML]{036400} -1.7052}          \\ \hline
\end{tabular}
}
\caption{\small \textit{{Generalization performance of individual fairness and detection utility on different backbones trained on FF++. We report both intra-domain (FF++) results and cross-domain (\ie, DFDC, Celeb-DF, DFD) generalization performance of Ori, Naive, and Ours. } }}
\vspace{-7mm}
\label{tab:cross_domain}
\end{table*}

\begin{figure*}[t]
    \centering
    \includegraphics[width=1.0\linewidth]{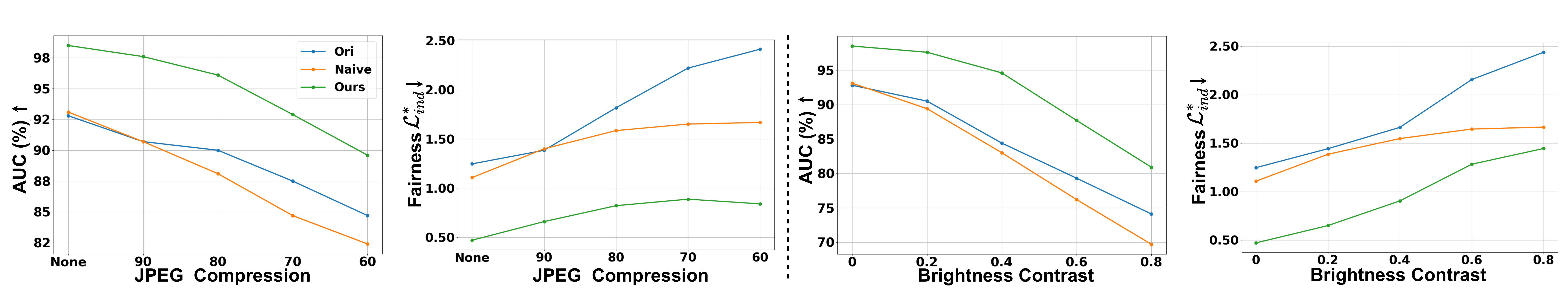}
    \vspace{-6mm}
    \caption{\small \textit{Robustness evaluation on FF++ under common post-processing methods.}}
    \label{fig:robustness}
\end{figure*}

\subsection{Experimental Settings}
\smallskip
\noindent
\textbf{Datasets}.
To comprehensively assess the proposed method, we utilize five widely-used public datasets: FaceForensics++ (FF++) \cite{rossler2019faceforensics++}, Deepfake Detection (DFD) \cite{googledeepfakes2019}, Deepfake Detection Challenge (DFDC) \cite{DFDC2020}, Celeb-DF \cite{li2020celebdf}, and the latest million-scale dataset, AI-Face~\cite{lin2024ai}. AI-Face notably includes face images generated by advanced diffusion models, which are absent from the other four datasets, enabling the evaluation of our method against more sophisticated generative techniques. 
For intra-domain evaluation, we train models on each dataset's training set and evaluate them on its corresponding test set. For cross-domain evaluation, following common practice in previous works~\cite{ju2023improving, lin2024preserving, Yan_2023_ICCV}, we train our model on FF++ and test it on the remaining datasets. Since we compare against existing fairness-enhanced methods~\cite{ju2023improving, lin2024preserving}, we adopt their data processing and annotation settings to ensure fair comparisons. Details of each dataset are provided in Appendix~\ref{appendix:datasets}. {We conduct experiments using three popular backbones: Xception~\cite{chollet2017xception}, ResNet-50~\cite{he2016deep}, and EfficientNet-B3~\cite{tan2019efficientnet}. Unless explicitly mentioned, all methods are employed on Xception~\cite{chollet2017xception} backbone.}

\smallskip
\noindent
\textbf{Evaluation Metrics}. For detection utility, we use the Area Under the Curve (AUC), consistent with prior works~\cite{ju2023improving, lin2024preserving, Yan_2023_ICCV}. To evaluate individual fairness, we treat $\mathcal{L}_{\text{ind}}^{naive}$ and $\mathcal{L}_{\text{ind}}^{*}$ as the traditional and proposed individual fairness metrics, respectively.

\smallskip
\noindent
\textbf{Baseline Methods}.
To clearly demonstrate our method's effectiveness in improving individual fairness in deepfake detection, we first define two fundamental baselines: (1) \textbf{Ori}, a backbone (\eg, Xception~\cite{chollet2017xception}) trained solely with standard CE loss, and (2) \textbf{Naive}, the Ori detector trained using CE loss with traditional individual fairness regularization $\mathcal{L}_{ind}^{naive}$. Additionally, to illustrate our method's plug-and-play flexibility, we integrate it into various state-of-the-art detectors across multiple categories, including spatial-based (UCF~\cite{Yan_2023_ICCV}, CORE~\cite{ni2022core}, RECCE~\cite{cao2022end}), frequency-based (F3Net~\cite{qian2020thinking}, SPSL~\cite{liu2021spatial}, SRM~\cite{luo2021generalizing}), and fairness-enhanced detectors (DAW-FDD~\cite{ju2023improving}, DAG-FDD~\cite{ju2023improving}, PG-FDD~\cite{lin2024preserving}).

\smallskip
\noindent
\textbf{Implementation Details}. All experiments are implemented in PyTorch and conducted using a NVIDIA A100 GPU. We train all models with a batch size of 32 for 60 epochs, using the SGD optimizer with a learning rate of $\beta=5 \times 10^{-4}$. In the anchor learning module, we utilize the entire training dataset as the reference set, and the masking schedule hyperparameter $\alpha$ is set to 0.2 empirically. In the semantic-agnostic individually fair learning module, we use a patch size of $P=32$. In the optimization module, the perturbation magnitude for SAM is set to $\gamma=0.05$. The trade-off hyperparameter $\lambda$ in the final learning objective Eq. (\ref{eq:final_loss}) is fixed to 0.001, which we experimentally found to yield the best performance across all datasets. The $\tau$ in Eq. (\ref{eq:base}) and Eq. (\ref{eq:ours_ind}) are fixed as 0.00005 for inference of all experiments. Additional details on $\lambda$ and $\tau$ selection are provided in Appendix~\ref{appendix:implementation_details}.

\subsection{Results}\label{subsec:results}
\smallskip
\noindent
\textbf{Performance on Intra-domain Testing}. We first evaluate the performance of our method under the intra-domain setting, where each model is trained and tested on the same dataset. As shown in Table~\ref{tab:in_domain}, our method consistently outperforms both Ori and Naive across all datasets in terms of individual fairness. For example, on the FF++ dataset, our method improves the AUC by 5.77\% over Ori and reduces the individual fairness loss $\mathcal{L}_{ind}^{*}$ by 0.7763. Compared to Naive, it achieves a 5.41\% improvement in AUC and a 0.6373 reduction in $\mathcal{L}_{ind}^{*}$.
These consistent reductions in individual fairness across all five datasets—including AI-Face~\cite{lin2024ai}, which contains diffusion model-generated faces—demonstrate the effectiveness of our method in enhancing individual fairness.

\smallskip
\noindent
\textbf{Performance of Fairness Generalization}. To assess the generalizability of our method, we conduct cross-domain evaluations where all models are trained on FF++ and tested on other datasets. As shown in Table~\ref{tab:cross_domain}, it is clear that our proposed approach consistently yields improvements in both detection utility and individual fairness. For instance, using the Xception backbone, our method achieves a 3.25\% increase in the AUC compared to Ori and a 4.35\% improvement over Naive on DFDC, while reducing the individual fairness loss, $\mathcal{L}_{ind}^{*}$, by 2.1235 and 1.1223, respectively. Similarly, on Celeb-DF, our method improves the AUC by 13.97\% over Ori and 4.86\% over Naive, alongside reductions in $\mathcal{L}_{ind}^{*}$ of 0.4539 and 0.5911, respectively. All these observations clearly demonstrate the importance of breaking the semantic correlation to improve individual fairness, as well as leveraging anchor learning to enhance the generalization of individual fairness and detection utility. By enforcing prediction consistency among samples with similar manipulation characteristics, our model learns to prioritize forgery cues over semantic identity, which leads to the observed improvements in performance.

\smallskip
\noindent
\textbf{Individual Fairness Adaptability of Different Backbones}. To evaluate the adaptability of our method across different model architectures, we conduct experiments using three popular backbones: Xception~\cite{chollet2017xception}, ResNet-50~\cite{he2016deep}, and EfficientNet-B3~\cite{tan2019efficientnet}. The results in Table~\ref{tab:cross_domain} indicate that our method based on different backbones demonstrates similar superior results. Such outcomes suggest that our approach is not tied to a specific architecture, but is broadly effective and applicable across diverse backbone designs.

\smallskip
\noindent
\textbf{Plug-and-Play Flexibility Across Different Detectors.} To validate the broad applicability of our framework, we integrate it into a wide range of state-of-the-art deepfake detectors across spatial-based, frequency-based, and fairness-enhanced categories.
Since anchor learning may not be compatible with some detectors due to architectural constraints (\eg, UCF~\cite{Yan_2023_ICCV} and PG-FDD~\cite{lin2024preserving}), we apply only the semantic-agnostic individually fair learning in these methods. 
As demonstrated in Table~\ref{tab:plug_play}, regardless of the different detection strategies, our method successfully and consistently improves individual fairness and detection utility. For example, when applied to SRM~\cite{luo2021generalizing}, a frequency-based model, our method reduces $\mathcal{L}_{ind}^*$ from 0.5797 to 0.3688 and improves AUC from 96.29\% to 97.06\%. Particularly notable is the improvement on PG-FDD~\cite{lin2024preserving}, where $\mathcal{L}_{ind}^*$ is lowered by 0.1387, demonstrating the compatibility of our framework with existing fairness-focused solutions. These results confirm that our method is model-agnostic and can be easily integrated to enhance individual fairness across diverse deepfake detection strategies.

\smallskip
\noindent
\textbf{Performance of Fairness Robustness.} To analyze the robustness of the proposed method, different types of perturbations are applied to the test set of FF++. Fig.~\ref{fig:robustness} illustrates the trends in both detection utility and individual fairness under different levels of JPEG Compression~\cite{cozzolino2023raising} and Brightness Contrast~\cite{yan2023deepfakebench}. Additional results on other post-processing operations are provided in Appendix~\ref{appendix:addtional_results}. Overall, these perturbations tend to wash out forensic traces, leading to evident performance degradation in detectors. Despite this, our method demonstrates remarkable resilience, with only a modest increase in fairness loss and minimal degradation in detection utility. This robustness can be attributed to the semantic-agnostic individually fair learning module, which effectively suppresses high-level identity semantics and exposes manipulation artifacts that remain informative even under visual distortions. 

\begin{table}[t!]
\centering
\scalebox{0.85}{
\begin{tabular}{c|c|ccc}
\hline
                                                                               &                                        & Utility(\%)\textuparrow                            & \multicolumn{2}{c}{Fairness\textdownarrow}                                                      \\ \cline{3-5} 
\multirow{-2}{*}{Type}                                                         & \multirow{-2}{*}{Method}               & AUC                                    & $\mathcal{L}_{ind}^{naive}$        & $\mathcal{L}_{ind}^{*}$        \\ \hline \hline
                                                                               & UCF~\cite{Yan_2023_ICCV}                                    & 97.27                                  & 1.1040                                  & 0.8617                                  \\
                                                                               & \cellcolor[HTML]{EFEFEF}Ours (UCF)     & \cellcolor[HTML]{EFEFEF}\textbf{97.44} & \cellcolor[HTML]{EFEFEF}\textbf{0.8617} & \cellcolor[HTML]{EFEFEF}\textbf{0.6366} \\
                                                                               & CORE~\cite{ni2022core}                                   & 96.11                                  &0.7020                                         & 0.4326                                  \\
                                                                               & \cellcolor[HTML]{EFEFEF}Ours (CORE)     & \cellcolor[HTML]{EFEFEF}\textbf{97.06} & \cellcolor[HTML]{EFEFEF}\textbf{0.5294}                & \cellcolor[HTML]{EFEFEF}\textbf{0.2900} \\
                                                                               & RECCE~\cite{cao2022end}                                 &96.49                                        &0.6082                                         & 0.3504                                        \\
\multirow{-6}{*}{Spatial}                                                      & \cellcolor[HTML]{EFEFEF}Ours (RECCE)   & \cellcolor[HTML]{EFEFEF}\textbf{97.12}               & \cellcolor[HTML]{EFEFEF}\textbf{0.5128}               & \cellcolor[HTML]{EFEFEF}\textbf{0.2950}                \\ \hline
                                                                               & F3Net~\cite{qian2020thinking}                                  & 96.76                                  & 1.4364                                  & 1.0940                                  \\
                                                                               & \cellcolor[HTML]{EFEFEF}Ours (F3Net)    & \cellcolor[HTML]{EFEFEF}\textbf{97.60} & \cellcolor[HTML]{EFEFEF}\textbf{0.8612} & \cellcolor[HTML]{EFEFEF}\textbf{0.6047} \\
                                                                               & SPSL~\cite{liu2021spatial}                                   & 96.64                                  & 1.9358                                  & 1.5563                                  \\
                                                                               & \cellcolor[HTML]{EFEFEF}Ours (SPSL)     & \cellcolor[HTML]{EFEFEF}\textbf{97.31} & \cellcolor[HTML]{EFEFEF}\textbf{1.1792} & \cellcolor[HTML]{EFEFEF}\textbf{0.8418} \\
                                                                               & SRM~\cite{luo2021generalizing}                                    & 96.29                                  & 0.9203                                  & 0.5797                                  \\
\multirow{-6}{*}{Frequency}                                                    & \cellcolor[HTML]{EFEFEF}Ours (SRM)      & \cellcolor[HTML]{EFEFEF}\textbf{97.06} & \cellcolor[HTML]{EFEFEF}\textbf{0.6867} & \cellcolor[HTML]{EFEFEF}\textbf{0.3688} \\ \hline
                                                                               & DAG-FDD~\cite{ju2023improving}                                & 97.13                                  & 1.1014                                  & 0.8634                                  \\
                                                                               & \cellcolor[HTML]{EFEFEF}Ours (DAG-FDD) & \cellcolor[HTML]{EFEFEF}\textbf{97.33}               & \cellcolor[HTML]{EFEFEF}\textbf{0.9371}                & \cellcolor[HTML]{EFEFEF}\textbf{0.6918}                \\
                                                                               & DAW-FDD~\cite{ju2023improving}                                & 97.46                                  & 1.0150                                  & 0.7195                                  \\
                                                                               & \cellcolor[HTML]{EFEFEF}Ours (DAW-FDD) & \cellcolor[HTML]{EFEFEF}\textbf{97.66}               & \cellcolor[HTML]{EFEFEF}\textbf{0.9933}               & \cellcolor[HTML]{EFEFEF}\textbf{0.6995}                 \\
                                                                               & PG-FDD~\cite{lin2024preserving}                                 & 97.59                                  & 0.8538                                  & 0.6063                                  \\
\multirow{-6}{*}{\begin{tabular}[c]{@{}c@{}}Fairness-\\ enhanced\end{tabular}} & \cellcolor[HTML]{EFEFEF}Ours (PG-FDD)  & \cellcolor[HTML]{EFEFEF}\textbf{97.93} & \cellcolor[HTML]{EFEFEF}\textbf{0.7022} & \cellcolor[HTML]{EFEFEF}\textbf{0.4676} \\ \hline
\end{tabular}
}
\caption{\small \textit{{Flexibility and adaptability evaluation of our method by integrating it with different state-of-the-art detection methods.} }}
\vspace{-3mm}
\label{tab:plug_play}
\end{table}

\begin{table}[t]
\centering
\scalebox{0.8}{
\begin{tabular}{cll|ccc}
\hline
\multicolumn{3}{c|}{}                                                                                    & Utility(\%)\textuparrow        & \multicolumn{2}{c}{Fairness\textdownarrow}                                        \\ \cline{4-6} 
\multicolumn{3}{c|}{\multirow{-2}{*}{Method}}                                                            & AUC            & $\mathcal{L}_{ind}^{naive}$ & $\mathcal{L}_{ind}^{*}$ \\ \hline \hline
\multicolumn{3}{c|}{Residual$\rightarrow$\textcolor{blue}{Frequency}$\rightarrow$\textcolor{red}{Patch}}                                & 98.39          & 1.2701                           & 1.0275                           \\
\multicolumn{3}{c|}{Residual$\rightarrow$\textcolor{red}{Patch}$\rightarrow$\textcolor{blue}{Frequency}}                                & 98.27          & 1.0195                           & 0.7867                           \\
\multicolumn{3}{c|}{\textcolor{red}{Patch}$\rightarrow$\textcolor{blue}{Frequency}$\rightarrow$Residual}                                & 98.27          & 1.0095                           & 0.7775                           \\
\multicolumn{3}{c|}{\textcolor{blue}{Frequency}$\rightarrow$\textcolor{red}{Patch}$\rightarrow$Residual}                                & \textbf{98.58} & 0.9197                           & 0.5726                           \\
\multicolumn{3}{c|}{\textcolor{blue}{Frequency}$\rightarrow$Residual$\rightarrow$\textcolor{red}{Patch}}                                & 98.44          & 0.8794                           & 0.5291                           \\
\rowcolor[HTML]{EFEFEF} 
\multicolumn{3}{c|}{\cellcolor[HTML]{EFEFEF}\textcolor{red}{Patch}$\rightarrow$Residual$\rightarrow$\textcolor{blue}{Frequency} (Ours)} & 98.53          & \textbf{0.8153}                  & \textbf{0.4709}                  \\ \hline
\end{tabular}
}
\caption{\small \textit{{Ablation study on the processing order of each component in the semantic-agnostic individually fair learning.} }}
\vspace{-8mm}
\label{tab:processing_order}
\end{table}

\begin{table}[t!]
\centering
\scalebox{0.70}{
\begin{tabular}{ccccc|ccc}
\hline
\multicolumn{5}{c|}{Method}                      & Utility(\%)\textuparrow    & \multicolumn{2}{c}{Fairness\textdownarrow}                                        \\ \hline
SAM   & Anchoring & Residual & Patch & Frequency & AUC            & $\mathcal{L}_{ind}^{naive}$ & $\mathcal{L}_{ind}^{*}$ \\ \hline \hline
 & \checkmark      & \checkmark     & \checkmark  & \checkmark      & 97.88          & 1.1711                           & 0.8435 ({\color[HTML]{3531ff}{0.3726}})                   \\
\checkmark  &      & \checkmark     & \checkmark  & \checkmark      & 97.97          & 1.5219                           & 1.0667 ({\color[HTML]{3531ff} 0.5958})                   \\
\checkmark  & \checkmark      &     & \checkmark  & \checkmark      & 98.12          & 1.1452                           & 0.9072 ({\color[HTML]{3531ff} 0.4363})                   \\
\checkmark  & \checkmark      & \checkmark     &  & \checkmark      & 98.03          & 1.2551                           & 1.0126 ({\color[HTML]{3531ff} 0.5417})                   \\
\checkmark  & \checkmark      & \checkmark     & \checkmark  &      & 97.88          & 1.3244                           & 1.0801 ({\color[HTML]{3531ff} 0.6091})                   \\
\rowcolor[HTML]{EFEFEF} 
\checkmark  & \checkmark      & \checkmark     & \checkmark  & \checkmark      & \textbf{98.53} & \textbf{0.8153}                  & \textbf{0.4709}                  \\ \hline
\end{tabular}
}
\caption{\small \textit{{Ablation of the impact of each component in our framework. The ({\color[HTML]{3531ff}{blue}}) highlights the difference between each method and our method (last row), reflecting the impact of removing each component from the full method.} }}
\vspace{-6mm}
\label{tab:component_impact}
\end{table}

\subsection{Ablation Study}

\smallskip
\noindent
\textbf{Effectiveness of Semantic-Agnostic Processing Order.} Our semantic-agnostic individually fair learning module is designed to suppress semantic information and emphasize manipulation artifacts by applying three key operations: patch shuffle, denoising (with residual extraction), and frequency transformation. To validate the necessity and optimal ordering of these components, we conduct an ablation study by permuting their execution order and evaluating their impact on both detection utility and individual fairness. As observed in Table~\ref{tab:processing_order}, executing the operations in the order $\textbf{\textcolor{red}{Patch}}\rightarrow\textbf{Residual}\rightarrow\textbf{\textcolor{blue}{Frequency}}$ -- achieves the best overall performance, with the lowest individual fairness $\mathcal{L}_{ind}^{*}$ of 0.4709 and strongest AUC of 98.53\%. 
Other orderings lead to suboptimal fairness. For instance, applying residual extraction before patch shuffling (\eg, $\textbf{Residual}\rightarrow\textbf{\textcolor{red}{Patch}}\rightarrow\textbf{\textcolor{blue}{Frequency}}$) is less effective because semantic information is already attenuated in the residual, limiting the effectiveness of subsequent patch shuffling. Likewise, applying frequency transformation too early (\eg, $\textbf{\textcolor{blue}{Frequency}}\rightarrow\textbf{\textcolor{red}{Patch}}\rightarrow\textbf{Residual}$) is suboptimal because patch shuffling in the frequency space cannot effectively disrupt spatial semantics, making it less capable of suppressing identity-related cues. These observations confirm that our proposed order—$\textbf{\textcolor{red}{Patch}}\rightarrow\textbf{Residual}\rightarrow\textbf{\textcolor{blue}{Frequency}}$—is essential for maximizing semantic suppression and enhancing individual fairness.

\begin{figure}[t]
    \centering
    \includegraphics[width=1\linewidth]{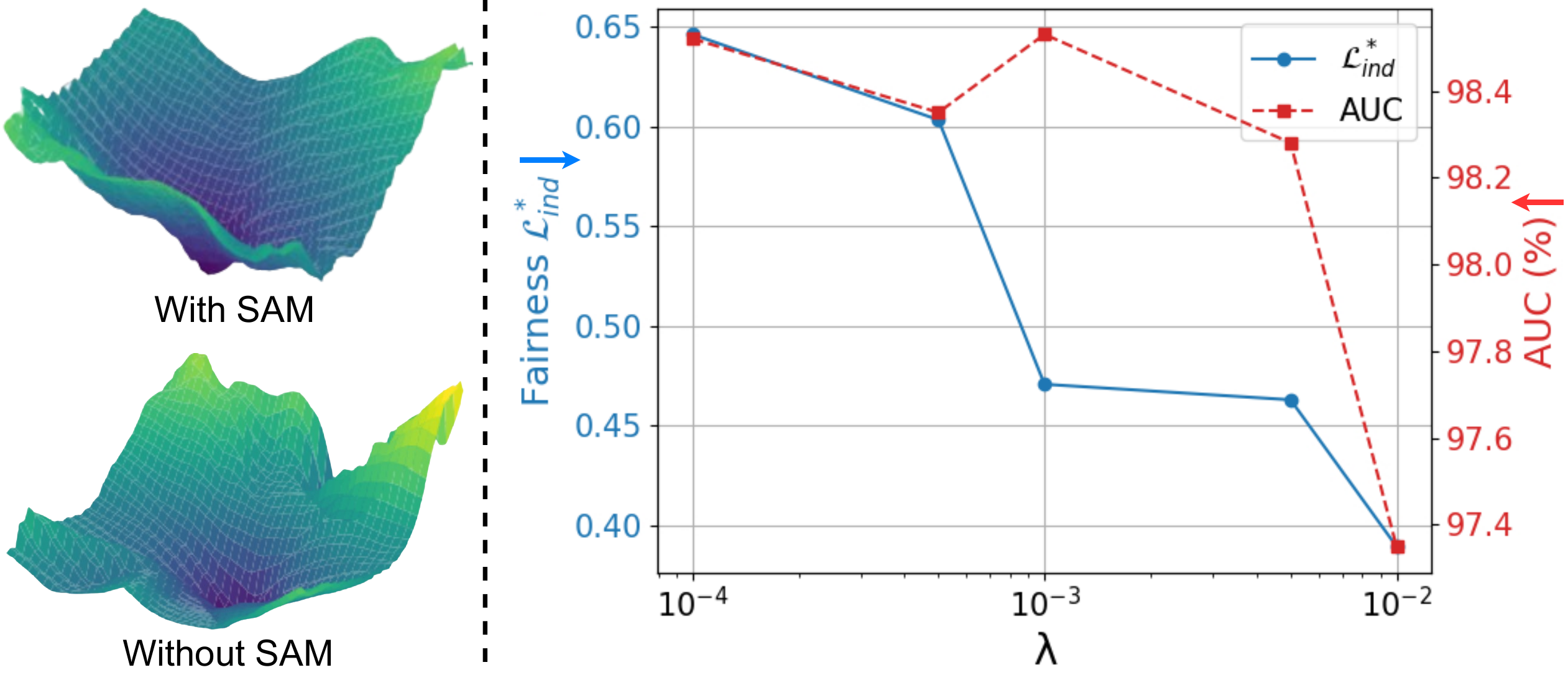}
    \vspace{-6mm}
    \caption{\small \textit{ (Left) The the loss landscape with SAM and w/o SAM. (Right) Sensitivity analysis of the trade-off hyperparameter $\lambda$.}}
    \label{fig:sensitivity}
    \vspace{-4mm}
\end{figure}

\smallskip
\noindent
\textbf{Effects of Each Component.} To assess the contribution of each component within our proposed framework, we conduct an ablation study by gradually removing each component. The impact of each component is highlighted in blue in Table~\ref{tab:component_impact}, where the differences are measured relative to the full method (last row). Among all components, frequency transformation shows the greatest contribution, reducing the individual fairness loss $\mathcal{L}_{ind}^*$ by 0.6091. Based on the differences in $\mathcal{L}_{ind}^*$, we rank the importance of components as follows: frequency transformation > anchor learning > patch shuffle > denoising and residual extraction > SAM. These results confirm that each module meaningfully contributes to improving individual fairness, with frequency-domain analysis and reference-based anchoring playing particularly critical roles. Additionally. Fig.~\ref{fig:sensitivity} (Left) visually illustrates the effectiveness of SAM~\cite{foret2020sharpness} in flattening the loss landscape, which reduces the model's sensitivity to perturbations and consequently improves fairness generalization.

\smallskip
\noindent
\textbf{Sensitivity Analysis.} To investigate the influence of the trade-off hyperparameter $\lambda$ in our final objective~\ref{eq:final_loss}, we conduct a sensitivity analysis by varying $\lambda\in \{0.0001, 0.0005, 0.001, 0.005, 0.01\}$ while keeping $\tau$ constant at 0.00001. As shown in Fig.~\ref{fig:sensitivity} (Right), there is a clear trade-off between detection utility and fairness. Increasing $\lambda$ leads to a consistent improvement in fairness, but at the cost of reduced utility. Notably, $\lambda = 0.001$ achieves a favorable balance between utility and fairness, maintaining high AUC while lowering the fairness loss.
Therefore, $\lambda = 0.001$ is set as the default value in all our experiments.


\section{Conclusion}
In this paper, we challenge the conventional understanding of individual fairness in the context of deepfake detection. We are the first to identify a fundamental contradiction: the original principle of individual fairness fails in deepfake detection due to the high semantic similarity between real and fake samples. To address this issue, we propose a novel and generalizable framework that integrates seamlessly with existing detectors to improve individual fairness without sacrificing detection utility.

\smallskip
\noindent
\textbf{Limitation}.  Although the framework is designed to be generalizable across various detectors, its integration may require additional tuning of hyperparameter $\lambda$ depending on the specific model architecture and dataset characteristics. While our paper currently lacks a formal theoretical foundation, we plan to address this in future works. Nonetheless, our paper does include formal analyses on the challenges of applying traditional individual fairness to deepfake detection as well as the advantages of our proposed approach. 

\smallskip
\noindent
\textbf{Future Work}. Although the AI-Face dataset is used in our experiments because it contains deepfake video-based images exhibiting semantic similarity issues, the effectiveness of our method on detecting faces generated solely by GANs or diffusion models remains unclear. These generated faces typically differ significantly from real ones and may not exhibit the semantic similarity challenges discussed in this work. How do we improve individual fairness in such a scenario? We plan to investigate this problem in future research. We also plan to extend our methods to improve individual fairness in detecting AI-generated voices. 

\begin{acks}
We thank anonymous reviewers for constructive comments. This work is supported by the U.S. National Science Foundation (NSF) under grant IIS-2434967 and the National Artificial Intelligence Research Resource (NAIRR) Pilot and TACC Lonestar6. The views, opinions and/or fndings expressed are those of the author and should not be interpreted as representing the offcial views or policies of NSF and NAIRR Pilot.
\end{acks}

\bibliographystyle{ACM-Reference-Format}
\balance
\bibliography{sample-base}

\newpage
\clearpage
\pagenumbering{arabic}
\title{Supplementary Materials: Rethinking Individual Fairness in Deepfake Detection}

\maketitle

\section{Related Work}
\textbf{Deepfake Detection.}
Current deepfake detection methods can be categorized into three primary groups based on the features they employ. The first category hinges on identifying inconsistencies in the \textit{physical and physiological} characteristics of deepfakes~\cite{lin2024detecting}. For example, inconsistent corneal specular highlights \cite{hu2021exposing}, the irregularity of pupil shapes \cite{guo2022eyes,guo2022open}, eye blinking patterns \cite{li2018ictu}, eye color difference \cite{matern2019exploiting}, facial landmark locations \cite{yang2019exposing}, etc. 
The second category concentrates on \textit{signal-level} artifacts introduced during the synthesis process, especially those from the frequency domain \cite{qian2020thinking}. 
These methods encompass various techniques, such as examining disparities in the frequency spectrum \cite{khayatkhoei2022spatial, dzanic2020fourier} and utilizing checkerboard artifacts introduced by the transposed convolutional operator \cite{frank2020leveraging, zhang2019detecting}.
However, the methods in the above two categories usually exhibit relatively low detection performance.
Therefore, the largest portion of existing detection methods fall into the \textit{data-driven} category, including \cite{guo2022robust, pu2022learning, Yan_2023_ICCV,liu2024forgery,hong2024contrastive,choi2024exploiting}. These methods leverage various types of DNNs trained on both authentic and deepfake videos to capture specific discernible artifacts. 
\textit{While these methods achieved strong detection utility, they overlooked detection fairness, resulting in large performance disparities among groups (\eg, gender)}. 

\smallskip
\noindent
\textbf{Fairness in Deepfake Detection.}
Many studies \cite{trinh2021examination,hazirbas2021towards,pu2022fairness,xu2024analyzing,lin2024ai,nadimpalli2022gbdf} have identified fairness issues in deepfake detection. For instance, Trinh et al. \cite{trinh2021examination} found biases present in both datasets and detection models, resulting in significant disparities in error rates across demographic groups. Similar findings were reported by Hazirbas et al. \cite{hazirbas2021towards}. Pu et al. \cite{pu2022fairness} evaluated the fairness of the MesoInception-4 model using the FF++ dataset, observing performance bias related to gender. 
Nadimpalli et al. \cite{nadimpalli2022gbdf} also highlighted substantial biases and proposed a preprocessing method involving a gender-balanced dataset to mitigate gender-based performance disparities. However, this approach delivered only modest improvements and necessitated extensive data annotation efforts. To further support fairness-oriented algorithm development, \cite{xu2024analyzing} and  \cite{lin2024ai} performed comprehensive analyses of bias in deepfake detection, augmenting datasets with diverse demographic annotations. Recent works by Ju et al. \cite{ju2023improving} and Lin et al. \cite{lin2024preserving} have subsequently concentrated on enhancing group-level fairness through algorithmic interventions. However, these studies have not comprehensively examined fairness at the individual level with solutions, an aspect that \textit{remains largely unaddressed in recent deepfake detection research and constitutes the primary focus of our work}.

\section{Algorithm}
Algorithm \ref{alg:joint_optimization} is the algorithm of optimization with individual fairness, which integrates a loss flattening strategy based on sharpness-aware minimization~\cite{foret2020sharpness}, and is implemented throughout the end-to-end training process.

\begin{algorithm}[h]
\caption{Optimization with Individual Fairness}\label{alg:joint_optimization} \footnotesize
\begin{algorithmic}[1]  
\State \textbf{Input:} A training dataset $\mathcal{S}$, a reference set $\mathcal{R} \subseteq \mathcal{S}$, max\_iterations $T$, num\_batch, learning rate $\beta$, SAM perturbation parameter $\gamma > 0$, and mask probability $\alpha > 0$.
\textbf{Initialization:} $\theta_0$, $l=0$   
\For{$t = 1$ to T}
    \For{$b = 1$ to num\_batch}
            \State Sample a mini-batch $S_b=\{(X_i,Y_i)\}_{i=1}^{|S_b|}$ from $\mathcal{S}$
            \State Sample reference samples $R_b=\{\bar{r}_i\}_{i=1}^{|S_b|}$ from $\mathcal{R}$
            \State \noindent$\triangleright$ \textbf{Step 1: Anchor Learning} 
            \State Draw $Mask \in \{0,1\}$ from Bernoulli$(\alpha)$
            \State Compute $\mathcal{L}_{CE}^*$ based on Eq. (2)
            \State \noindent$\triangleright$ \textbf{Step 2: Semantic-Agnostic Individual Fair Learning}
            \State $X' \gets \text{Reconstruct}(\text{Shuffle}(\text{Partition}(X, P)))$, $\forall X \in \mathcal{S}_b$
            \State $\mathcal{F}(\overline{X}') \gets \text{DFT}(X' - D(X'))$
            \State Compute $\mathcal{L}_{ind}^*$ based on Eq.~(3)
            \State \noindent$\triangleright$
            \textbf{Step 3: Optimization}
            \State Compute $\epsilon^*$ based on Eq. (5)
            \State Compute gradient approximation for $
    \begin{aligned}      
    \min_\theta \mathcal{L}\textbf{(}\theta+\epsilon^*\textbf{)}
    \end{aligned}$
            \State Update $\theta$: $\theta_{l+1} \leftarrow \theta_{l} - \beta \nabla_\theta \mathcal{L}\big|_{\theta_l + \epsilon^*}$
            \State $l \leftarrow l + 1$ 
    \EndFor
\EndFor
\State \textbf{Output:} $\theta_{l^*}$, where $l^*$ is the best iterate satisfying $
    \begin{aligned}      
    \min_\theta \mathcal{L}\textbf{(}\theta+\epsilon^*\textbf{)}
    \end{aligned}$ with the lowest objective.
\end{algorithmic}
\end{algorithm}

\section{Additional Experimental Settings}
\subsection{Description of The Datasets}\label{appendix:datasets}
Here, we show the total number of training and test samples for each dataset included in our experiments, as in Table \ref{tab:dataset_splits}. While our method does not inherently require the demographic attributes (\ie,  gender, race) included in the datasets, we leveraged these annotations to benchmark our proposed method against fairness-aware baselines such as DAW-FDD \cite{ju2023improving} and PG-FDD \cite{lin2024preserving}, which explicitly incorporate these features into their pipelines. 
\begin{table}[t]
\centering
\scalebox{0.95}{
\begin{tabular}{ccc}
\toprule
Dataset & \multicolumn{1}{c}{Training} & \multicolumn{1}{c}{Test} \\
\midrule
FF++ \cite{rossler2019faceforensics++} & 76,139 & 25,401 \\
DFDC \cite{DFDC2020} & 71,478 & 22,857 \\
Celeb-DF \cite{li2020celebdf} & 93,595 & 28,458 \\
DFD \cite{googledeepfakes2019} & 23,834 & 9,385 \\
AI-Face \cite{lin2024ai} & 1,689,405 & 422,351 \\
\bottomrule
\end{tabular}
}
\caption{Number of training and test samples across datasets used in our experiments.}
\label{tab:dataset_splits}
\end{table}

\begin{figure*}[t]
    \centering
    \includegraphics[width=0.9\linewidth]{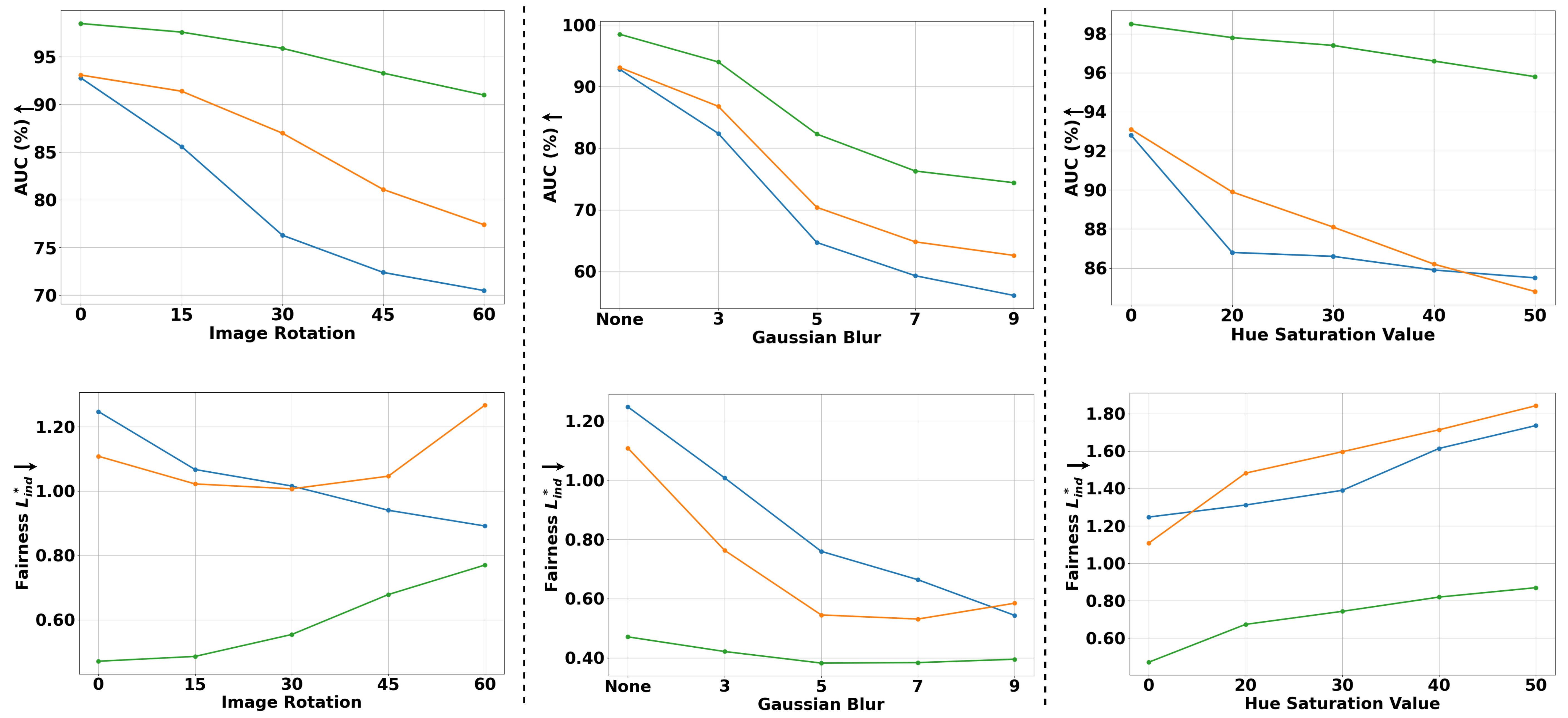}
    \vspace{-2mm}
    \caption{\small \textit{Robustness evaluation on FF++ under common post-processing methods.}}
    \label{fig:robustness-additional}
\end{figure*}

\subsection{The Formulation of Evaluation Metrics}
We have individual fairness metrics $\mathcal{L}^{naive}_{ind}$ and $\mathcal{L}^{*}_{ind}$ defined in Eq. (\ref{eq:base}) and Eq. (\ref{eq:ours_ind}), respectively. However, when evaluating $\mathcal{L}^{naive}_{ind}$ using our proposed method, which requires anchor learning, $\mathcal{L}^{naive}_{ind}$ is needed to be adapted to $\mathcal{L}^{naive '}_{ind}$. And when evaluating $\mathcal{L}^{*}_{ind}$ using the Naive and Ori method, $\mathcal{L}^{*}_{ind}$ is needed to be adapted to $\mathcal{L}^{*'}_{ind}$.

Specifically, our method leverages anchor training to improve individual fairness, which requires anchored input, so when evaluating $\mathcal{L}_{ind}^{naive}$, we adapt it by using the anchored inputs $X^a$ for the term $\spadesuit$, then the metric for the \textbf{evaluation} is formulated as:
\begin{equation*}
    \begin{aligned}
     & \mathcal{L}^{naive '}_{ind} = \sum_{i=1}^{n-1} \sum_{j=i+1}^{n} \big[|h(E(X_i^a)) - h(E(X_{j}^a))| - \tau\left\|X_i - X_{j}\right\|_2\big]_+.
    \end{aligned}
\end{equation*}

The Naive and Ori method only needs the original image as input, so when evaluating $\mathcal{L}^{*}_{ind}$, we adapt it by using the original inputs $X$ for the term $\spadesuit$, then the metric is formulated as:
\begin{equation*}
    \begin{aligned}
     & \mathcal{L}^{*'}_{ind} = \sum_{i=1}^{n-1} \sum_{j=i+1}^{n} \big[|h(E(X_i)) - h(E(X_{j}))| - \tau\|\mathcal{F}(\overline{X}'_i) - \mathcal{F}(\overline{X}'_{j})\|_2\big]_+.
    \end{aligned}
\end{equation*}

\subsection{Implementation Details}\label{appendix:implementation_details}
We experimentally found the optimal values of the trade-off hyperparameter $\lambda$ in the learning objective Eq. (\ref{eq:final_loss}) and $\tau$ in Eq. (\ref{eq:base}) and Eq. (\ref{eq:ours_ind}), as seen in Tables \ref{tab:intra}, \ref{tab:cross-hyperparameter}, and \ref{tab:all-hyperparameter}, that would yield the best results for both Naive and our methods. To achieve this, we performed a grid search over
\begin{enumerate}[noitemsep,topsep=0pt,leftmargin=*]
    \item $\lambda \in \{0.00001, 0.00005, 0.0001, 0.0005, 0.001, 0.005, 
 0.01, 0.1, 1.0\}$
 \item $\tau \in \{0.00001, 0.00005, 0.0001, 0.0005, 0.001, 0.01, 0.1\}$
\end{enumerate}
For all methods, we kept $\tau$ as 0.00005 for inference. 
\renewcommand{\arraystretch}{1.2}
\begin{table*}[t!]
\centering
\scalebox{0.84}{
\begin{tabular}{c|c|ccc|ccc|ccc|ccc}
\hline
                                  &                                       & \multicolumn{3}{c|}{FF++}                                                                                                  & \multicolumn{3}{c|}{DFDC}                                                                                                  & \multicolumn{3}{c|}{Celeb-DF}                                                                                              & \multicolumn{3}{c}{DFD}                                                                           \\ \cline{3-14} 
                                  &                                       & Utility(\%)\textuparrow                            & \multicolumn{2}{c|}{Fairness\textdownarrow}                                                     & Utility(\%)\textuparrow                            & \multicolumn{2}{c|}{Fairness\textdownarrow}                                                     & Utility(\%)\textuparrow                            & \multicolumn{2}{c|}{Fairness\textdownarrow}                                                     & Utility(\%)\textuparrow                            & \multicolumn{2}{c}{Fairness\textdownarrow}                                          \\ \cline{3-14} 
\multirow{-3}{*}{Type}        & \multirow{-3}{*}{Method}              & AUC                                    &$\mathcal{L}_{ind}^{naive}$               &$\mathcal{L}_{ind}^{*}$               & AUC                                    &$\mathcal{L}_{ind}^{naive}$               &$\mathcal{L}_{ind}^{*}$               & AUC                                    &$\mathcal{L}_{ind}^{naive}$               &$\mathcal{L}_{ind}^{*}$               & AUC                                    &$\mathcal{L}_{ind}^{naive}$               &$\mathcal{L}_{ind}^{*}$               \\ \hline \hline
\multirow{2}{*}{Spatial}                                   & UCF~\cite{Yan_2023_ICCV}                                   & 97.27                                  &  1.1040                                  & 0.8617                                  & 60.39                                  & 2.8065                                 & 2.3675                                  & 71.06                                  & 1.8744                                  & 1.5470                                  & 81.09                                  & 1.9635                                  & 1.6595          \\
                                  & \cellcolor[HTML]{EFEFEF}\textbf{Ours (UCF)} & \cellcolor[HTML]{EFEFEF}\textbf{97.44} & \cellcolor[HTML]{EFEFEF}\textbf{0.8617} & \cellcolor[HTML]{EFEFEF}\textbf{0.6366} & \cellcolor[HTML]{EFEFEF}\textbf{61.98} & \cellcolor[HTML]{EFEFEF}\textbf{2.4550} & \cellcolor[HTML]{EFEFEF}\textbf{2.0435} & \cellcolor[HTML]{EFEFEF}\textbf{74.59} & \cellcolor[HTML]{EFEFEF}\textbf{1.3544} & \cellcolor[HTML]{EFEFEF}\textbf{1.0527} & \cellcolor[HTML]{EFEFEF}\textbf{83.09} & \cellcolor[HTML]{EFEFEF}\textbf{1.7670} & \cellcolor[HTML]{EFEFEF}\textbf{1.4532} \\
       \hline
    \multirow{2}{*}{Frequency}                                  & SRM~\cite{luo2021generalizing}                                   & 96.29                                  & 0.9203                                  & 0.5797                                  & 59.65                                  & 0.4414                                  & 0.1546                                  & 72.67                                  & 0.7820                                  & 0.4252                                  & 80.16                                  & 0.7786                                  & 0.4357                                \\
                                  & \cellcolor[HTML]{EFEFEF}\textbf{Ours (SRM)} & \cellcolor[HTML]{EFEFEF}\textbf{97.06} & \cellcolor[HTML]{EFEFEF}\textbf{0.6867} & \cellcolor[HTML]{EFEFEF}\textbf{0.3688} & \cellcolor[HTML]{EFEFEF}\textbf{62.29} & \cellcolor[HTML]{EFEFEF}\textbf{0.3531} & \cellcolor[HTML]{EFEFEF}\textbf{0.1065} & \cellcolor[HTML]{EFEFEF}\textbf{75.22} & \cellcolor[HTML]{EFEFEF}\textbf{0.5234} & \cellcolor[HTML]{EFEFEF}\textbf{0.2039} & \cellcolor[HTML]{EFEFEF}\textbf{82.01} & \cellcolor[HTML]{EFEFEF}\textbf{0.5295} & \cellcolor[HTML]{EFEFEF}\textbf{0.2262} \\
   \hline
    \multirow{2}{*}{\shortstack{Fairness-\\enhanced}}                        & PG-FDD~\cite{lin2024preserving}                                   & 97.59                                  & 0.8538                                  & 0.6063                                  & 60.27                       & 2.1680                                  & 1.7380                                  & 72.82                                  & 1.8156                                  & 1.4588                                  & 81.91                                  & 1.3932                                  & 1.0854                                 \\
                                  & \cellcolor[HTML]{EFEFEF}\textbf{Ours (PG-FDD)} & \cellcolor[HTML]{EFEFEF}\textbf{97.93} & \multicolumn{1}{r}{\cellcolor[HTML]{EFEFEF}\textbf{0.7022}} & \cellcolor[HTML]{EFEFEF}\textbf{0.4676} & \cellcolor[HTML]{EFEFEF}\textbf{61.28}          & \cellcolor[HTML]{EFEFEF}\textbf{1.4271} & \cellcolor[HTML]{EFEFEF}\textbf{1.0366} & \cellcolor[HTML]{EFEFEF}\textbf{75.64} & \cellcolor[HTML]{EFEFEF}\textbf{0.9235} & \cellcolor[HTML]{EFEFEF}\textbf{0.6308} & \cellcolor[HTML]{EFEFEF}\textbf{83.46} & \cellcolor[HTML]{EFEFEF}\textbf{1.1473} & \cellcolor[HTML]{EFEFEF}\textbf{0.8702} \\
\hline
    \multirow{2}{*}{\shortstack{Transformer-\\based}}                              & ViT-B16 ~\cite{dosovitskiy2021image}                                  & 88.80                                  & 1.8851                                  & 1.50888                                  & 60.41                       & 1.9812                                  & 1.5934                                  & 79.73                                  & 1.6452                                  & 1.2792                                  & 73.03                                  & 1.9154                                  & 1.5267                                 \\
                                  & \cellcolor[HTML]{EFEFEF}\textbf{Ours (ViT-B16)} & \cellcolor[HTML]{EFEFEF}\textbf{90.00} & \multicolumn{1}{r}{\cellcolor[HTML]{EFEFEF}\textbf{1.6928}} & \cellcolor[HTML]{EFEFEF}\textbf{1.32843} & \cellcolor[HTML]{EFEFEF}\textbf{61.71}          & \cellcolor[HTML]{EFEFEF}\textbf{1.8799} & \cellcolor[HTML]{EFEFEF}\textbf{1.4911} & \cellcolor[HTML]{EFEFEF}\textbf{81.39} & \cellcolor[HTML]{EFEFEF}\textbf{1.4001} & \cellcolor[HTML]{EFEFEF}\textbf{1.0391} & \cellcolor[HTML]{EFEFEF}\textbf{74.36} & \cellcolor[HTML]{EFEFEF}\textbf{1.5569} & \cellcolor[HTML]{EFEFEF}\textbf{1.1756} \\

\end{tabular}
}
\caption{\small \textit{{Flexibility and adaptability of our method by integrating into different state-of-art detection methods. We train the models on FF++ and report the intra-domain (FF++) and cross-domain (DFDC, Celeb-DF, DFD) generalization performance. } }}
\vspace{-4mm}
\label{tab:plug_play_cross_domain}
\end{table*}
\begin{table*}[t]
\centering
\scalebox{0.9}{
\begin{tabular}{c|cccccccccc|c}
\hline
\multirow{3}{*}{Method} & \multicolumn{10}{c|}{Training}                                                                                                                                                                                                                                                                                                                            & \multirow{2}{*}{Testing} \\ \cline{2-11}
                        & \multicolumn{2}{c|}{FF++ \cite{rossler2019faceforensics++}} & \multicolumn{2}{c|}{DFDC \cite{DFDC2020}} & \multicolumn{2}{c|}{Celebdf-DF \cite{li2020celebdf}} & \multicolumn{2}{c|}{DFD \cite{googledeepfakes2019}} & \multicolumn{2}{c|}{AI-Face \cite{lin2024ai}} &                          \\ \cline{2-12} 
                        & $\lambda$                   & \multicolumn{1}{c|}{$\tau$}                    & $\lambda$          & \multicolumn{1}{c|}{$\tau$}           & $\lambda$                & \multicolumn{1}{c|}{$\tau$}                & $\lambda$               & \multicolumn{1}{c|}{$\tau$}                & $\lambda$                       & $\tau$                       & $\tau$                   \\ \hline
Ori                     & -                           & \multicolumn{1}{c|}{-}                         & -                  & \multicolumn{1}{c|}{-}                & -                        & \multicolumn{1}{c|}{-}                     & -                       & \multicolumn{1}{c|}{-}                     & -                               & -                            & 0.00005                  \\
Naive                   & 0.005                       & \multicolumn{1}{c|}{0.0001}                    & 0.005              & \multicolumn{1}{c|}{0.001}            & 0.005                    & \multicolumn{1}{c|}{0.001}                 & 0.005                   & \multicolumn{1}{c|}{0.001}                 & 0.0005                          & 0.001                        & 0.00005                  \\
Ours                    & 0.001                       & \multicolumn{1}{c|}{0.00001}                   & 0.001              & \multicolumn{1}{c|}{0.00001}          & 0.001                    & \multicolumn{1}{c|}{0.00001}               & 0.001                   & \multicolumn{1}{c|}{0.00001}               & 0.001                           & 0.00001                      & 0.00005                  \\ \hline
\end{tabular}
}
\caption{Optimum hyperparameters $\lambda$ and $\tau$ for the intra-domain experiments across five datasets.}
\label{tab:intra}
\end{table*}

\begin{table}[t]
\scalebox{0.93}{
\begin{tabular}{c|c|cc|l}
\hline
\multirow{2}{*}{Backbone}        & \multirow{2}{*}{Method} & \multicolumn{2}{c|}{Training} & \multicolumn{1}{c}{Testing} \\ \cline{3-5} 
                                 &                         & $\lambda$      & $\tau$       & \multicolumn{1}{c}{$\tau$}  \\ \hline
\multirow{3}{*}{Xception}        & Ori                     & -              & -            & 0.00005                     \\
                                 & Naive                   & 0.005          & 0.0001       & 0.00005                     \\
                                 & Ours                    & 0.001          & 0.00001      & 0.00005                     \\ \hline
\multirow{3}{*}{ResNet-50}       & Ori                     & -              & -            & 0.00005                     \\
                                 & Naive                   & 0.00005        & 0.001        & 0.00005                     \\
                                 & Ours                    & 0.001          & 0.00001      & 0.00005                     \\ \hline
\multirow{3}{*}{EfficientNet-B3} & Ori                     & -              & -            & 0.00005                     \\
                                 & Naive                   & 0.0005         & 0.001        & 0.00005                     \\
                                 & Ours                    & 0.001          & 0.00001      & 0.00005                     \\ \hline
\end{tabular}
}
\caption{Optimum hyperparameters $\lambda$ and $\tau$ used for generalization experiments across three backbone architectures trained on FF++.}
\label{tab:cross-hyperparameter}
\end{table}

\begin{table}[]
\scalebox{0.93}{
\begin{tabular}{c|c|cc|c}
\hline
\multirow{2}{*}{Type}                                                         & \multirow{2}{*}{Method}                          & \multicolumn{2}{c|}{Training}            & Testing \\ \cline{3-5} 
                                                                              &                                                  & \multicolumn{1}{c|}{$\lambda$} & $\tau$  & $\tau$  \\ \hline
\multirow{6}{*}{Spatial}                                                      & UCF \cite{Yan_2023_ICCV}      & \multicolumn{1}{c|}{-}         & -       & 0.00005 \\
                                                                              & Ours (UCF)                                       & \multicolumn{1}{c|}{0.001}     & 0.00001 & 0.00005 \\ \cline{2-5} 
                                                                              & CORE \cite{ni2022core}          & \multicolumn{1}{c|}{-}         & -       & 0.00005 \\
                                                                              & Ours (CORE)                                      & \multicolumn{1}{c|}{0.001}     & 0.00001 & 0.00005 \\ \cline{2-5} 
                                                                              & RECCE \cite{cao2022end}         & \multicolumn{1}{c|}{-}         & -       & 0.00005 \\
                                                                              & Ours (RECCE)                                     & \multicolumn{1}{c|}{0.001}     & 0.00001 & 0.00005 \\ \hline
\multirow{6}{*}{Frequency}                                                    & F3Net \cite{qian2020thinking}   & \multicolumn{1}{c|}{-}         & -       & 0.00005 \\
                                                                              & Ours (F3Net)                                     & \multicolumn{1}{c|}{0.001}     & 0.00001 & 0.00005 \\ \cline{2-5} 
                                                                              & SPSL \cite{liu2021spatial}      & \multicolumn{1}{c|}{-}         & -       & 0.00005 \\
                                                                              & Ours (SPSL)                                      & \multicolumn{1}{c|}{0.001}     & 0.00001 & 0.00005 \\ \cline{2-5} 
                                                                              & SRM \cite{luo2021generalizing}  & \multicolumn{1}{c|}{-}         & -       & 0.00005 \\
                                                                              & Ours (SRM)                                       & \multicolumn{1}{c|}{0.001}     & 0.00001 & 0.00005 \\ \hline
\multirow{6}{*}{\begin{tabular}[c]{@{}c@{}}Fairness-\\ enhanced\end{tabular}} & DAG-FDD \cite{ju2023improving}  & \multicolumn{1}{c|}{-}         & -       & 0.00005 \\
                                                                              & Ours (DAG-FDD)                                   & \multicolumn{1}{c|}{0.001}     & 0.00001 & 0.00005 \\ \cline{2-5} 
                                                                              & DAW-FDD \cite{ju2023improving}  & \multicolumn{1}{c|}{-}         & -       & 0.00005 \\
                                                                              & Ours (DAW-FDD)                                   & \multicolumn{1}{c|}{0.001}     & 0.00001 & 0.00005 \\ \cline{2-5} 
                                                                              & PG-FDD \cite{lin2024preserving} & \multicolumn{1}{c|}{-}         & -       & 0.00005 \\ 
                                                                              & Ours (PG-FDD)                                    & \multicolumn{1}{c|}{0.001}     & 0.00001 & 0.00005 \\ \hline
\end{tabular}
}
\caption{Optimum hyperparameters $\lambda$ and $\tau$ used for integration with different state-of-the-art detection methods.}
\label{tab:all-hyperparameter}
\end{table}

\section{Additional Experimental Results}
\subsection{Additional Robustness Experiments}\label{appendix:addtional_results}
In addition to the JPEG Compression~\cite{cozzolino2023raising} and Brightness Contrast~\cite{yan2023deepfakebench} methods employed for robustness evaluation in Section~\ref{subsec:results}, we present the results of three additional post-processing operations here: Rotation (RT)\cite{yan2023deepfakebench}, Hue Saturation Value (HSV)\cite{yan2023deepfakebench}, and Gaussian Blur (GB)~\cite{yan2023deepfakebench} shown in Figure \ref{fig:robustness-additional}.
Across all three post-processing methods, our method consistently outperforms both the Ori and the Naive baselines in terms of both utility and fairness. The results highlight our method’s ability to improve individual fairness even under challenging perturbations. This robustness is particularly evident in scenarios with high perturbation intensity, where our method sustains higher AUC and lower $\mathcal{L}_{\text{ind}}$ compared to the baselines. These findings complement the earlier robustness evaluations with JPEG Compression and Brightness Contrast, further validating the robustness and practical applicability of our proposed method in real-world deepfake detection scenarios, where images are usually subjected to post-processing corruptions. Here, we describe each post-processing method as follows:

\textbf{JPEG Compression}: Image compression introduces compression artifacts and reduces the image quality, simulating real-world scenarios where images may be of lower quality or have compression artifacts. 

\textbf{Gaussian Blur}: This post-processing reduces image detail and noise by smoothing it through averaging pixel values with a Gaussian kernel. 

\textbf{Hue Saturation Value}: Alters the hue, saturation, and value of the image within specified limits. This post-processing technique is used to simulate variations in color and lighting conditions. Adjusting the hue changes the overall color tone, saturation controls the intensity of colors, and value adjusts the brightness. 

\textbf{Random Brightness and Contrast}: This post-processing method adjusts the brightness and contrast of the image within specified limits. By applying random brightness and contrast variations, it introduces changes in the illumination and contrast levels of the images. This evaluates detector's robustness to different illumination conditions. 

\textbf{Rotation}: Rotates the image within a specified angle limit. This post-processing method is used to evaluate the detector's robustness to changes in the orientation of objects within the image.

\begin{table}[t]
\centering
\scalebox{0.8}{
\begin{tabular}{c|ccc}
\hline
                                   & Utility(\%)\textuparrow        & \multicolumn{2}{c}{Fairness\textdownarrow}      \\ \cline{2-4} 
\multirow{-2}{*}{Method}           & AUC            &  $\mathcal{L}_{ind}^{naive}$      &    $\mathcal{L}_{ind}^{*}$      \\ \hline \hline
Semantic-agnostic frequency for CE & 63.75          &    \textbf{0.4935}             & \textbf{0.1756}              \\
Anchored for distance              & 97.68          & 1.5250          & 1.0632          \\
\rowcolor[HTML]{EFEFEF} 
\textbf{Ours}                      & \textbf{98.53} & 0.8153 & 0.4709\\ \hline
\end{tabular}
}
\caption{Ablation study on the impact of the anchor learning
module and the semantic-agnostic individually fair learning
module. The first row applies semantic-agnostic preprocessing
 (patch shuffle, denoising, frequency transform) directly
to the input for classification, without using anchor learning.
The second row uses anchor learning for individual distance
calculation, but without semantic-agnostic preprocessing.} 
\label{tab:ablation_module}
\vspace{-4mm}
\end{table}

\subsection{Additional Generalization Performance}\label{appendix:additional_performance}
To further validate the robustness and applicability of our method, we integrated our semantic-agnostic individually fair learning module into a variety of state-of-the-art detectors, ranging from spatial- and frequency-based methods to fairness-enhanced and transformer-based frameworks, and evaluated their generalization ability across multiple datasets by training on FF++ and testing on others such as DFD and DFDC. As shown in Table~{\ref{tab:plug_play_cross_domain}}, our approach consistently improves both individual fairness and utility. For instance, when applied to UCF, our method reduced $\mathcal{L}_{ind}^*$ from 1.5470 to 1.0527 and increased the AUC from 71.06\% to 74.59\% on Celeb-DF. Similarly, when applied to ViT-B16, a Transformer-based model introduced in ~\cite{dosovitskiy2021image}, our method reduced $\mathcal{L}_{ind}^*$ from 1.5267 to 1.1756 and increased the AUC from 73.03\% to 74.36\% on DFD. These results demonstrate that our method can be effectively integrated into diverse deepfake detectors to enhance both individual fairness and utility across unseen domains. 

\subsection{Additional Ablation Study}\label{appendix:addtional_ablation}
\textbf{Effects of Each Module.} Table~\ref{tab:ablation_module} presents an ablation analysis to quantify the contributions of the anchor learning module and the semantic-agnostic individually fair learning module. When semantic-agnostic preprocessing is used for classification, the model fails to effectively learn meaningful features for classification, resulting in a significantly degraded detection performance (AUC = 63.75\%). As a result of this collapse, the individual fairness loss $\mathcal{L}_{ind}^*$ becomes uninformative. This highlights the importance of using anchor learning. When using anchored representations for distance calculation in individual fairness, the $\mathcal{L}_{ind}^*$ increased by 0.5923 compared to Ours. This demonstrates the necessity of semantic-agnostic preprocessing in improving individual fairness in deepfake detection.

\end{document}